\documentclass{article}
\usepackage{kr}

\usepackage{times}
\usepackage{soul}
\usepackage{url}
\usepackage[hidelinks]{hyperref}
\usepackage[utf8]{inputenc}
\usepackage[small]{caption}
\usepackage{graphicx}
\usepackage{amsmath}
\usepackage{amsthm}
\usepackage{booktabs}
\usepackage{algorithm}
\usepackage[noend]{algpseudocode}
\usepackage{algorithmicx,algpseudocode}

\usepackage{natbib}
\usepackage{graphicx}
\usepackage{latexsym}
\usepackage[dvipsnames]{xcolor}
\usepackage{amsmath,amsthm}
\usepackage[scr=boondox,  
            ]   
           {mathalpha}
\usepackage{tikz}
\usetikzlibrary{decorations.pathreplacing,calligraphy}
\tikzstyle{cnode}=[draw,circle,inner sep=1pt,minimum size=.5cm]

\usepackage{amssymb}
\usepackage{thm-restate}
\usepackage{etoolbox}
\usepackage{mathtools}

\DeclarePairedDelimiter\floor{\lfloor}{\rfloor}

\usepackage{listings}
\newcommand{\aspmodule}[1]{\Pi_{\mathit{#1}}}

\newcommand{\la}{\leftarrow}
\newcommand{\naf}{\textit{not}\, }

\lstdefinelanguage{asp}
{morekeywords={in,nb,ap,collider,var,dep,set,collider_desc,coll_desc,arrow,dpath,edge,dep,indep,rule,contrary,defeated_by_undefeated,out,derived,derivable,head,body,derived_from_undefeated,derivable_from_undefeated,applicable_rule,usable_by_in,assumption,defeated,usable_by_undefeated,att_by_undefeated,no_undef_closed,guess,in_derives,guess_derives,usable_by_guess,in_defeated_by_guess,target,guess_defeats}, 
literate={:-}{{$\la\ $}}1 {not}{{$\naf$}}1 {\\el}{{$\in$}}1 {\\cup}{{$\cup$}}1 {:~}{{$\law$}}1,
morecomment=[l]{\%},
}
\DeclareCaptionStyle{ruled}{labelfont=normalfont,labelsep=colon,strut=off}%
\floatstyle{ruled}
\newfloat{listing}{tb}{lst}{}
\floatname{listing}{Listing}          

\def\DAG{G}

\def\edgSet{\ensuremath{E}}

\def\nodeSet{\ensuremath{\mathbf{V}}}

\def\Xvar{\ensuremath{\mathbf{X}}}
\def\Yvar{\ensuremath{\mathbf{Y}}}
\def\Zvar{\ensuremath{\mathbf{Z}}}

\def\condSet{\ensuremath{\mathbf{Z}}}
\newcommand{\indep}{\!\perp \!\!\! \perp\!}
\newcommand{\notindep}{\not\!\perp \!\!\! \perp\!}
\newcommand{\given}{ \mid }
\def\nullH{\ensuremath{\mathcal{H}_0}}
\def\altH{\ensuremath{\mathcal{H}_0}}
\newcommand{\Gpath}{\ensuremath{\mathscr{p}}}
\newcommand{\tree}{\ensuremath{\mathscr{t}}}
\newcommand{\ctpath}{\ensuremath{\mathscr{p}_\mathscr{t}}}

\definecolor{cadmiumgreen}{rgb}{0.0, 0.75, 0.40}
\definecolor{richcarmine}{rgb}{0.85,0.00,0.23}

\usepackage{todonotes}

\def\mhaf{\ensuremath{\langle {\cal L}, \, {\cal R}, \, {\cal A},\, \overline{
\vrule height 5pt depth 3.5pt width 0pt
\hskip0.5em\kern0.4em}\rangle}}

\newtheorem{theorem}{Theorem}
\numberwithin{theorem}{section} 

\newtheorem{lemma}[theorem]{Lemma}
\newtheorem{proposition}[theorem]{Proposition}
\newtheorem{corollary}[theorem]{Corollary}

\newtheorem{definition}[theorem]{Definition}
\newtheorem{remark}[theorem]{Remark}

\newtheorem{example}[theorem]{Example}

\newcommand{\contraryempty}{\overline{\phantom{a}}}
\newcommand{\contrary}[1]{\overline{#1}}
\newcommand{\stable}{{\mathit{stb}}}
\newcommand{\stb}{\stable}
\newcommand{\adm}{\mathit{ad}}
\newcommand{\prf}{\mathit{pr}}
\newcommand{\pref}{\mathit{pr}}
\newcommand{\grd}{\mathit{gr}}
\newcommand{\com}{\mathit{co}}
\newcommand{\comp}{\mathit{co}}

\newcommand{\stage}{\mathit{stg}}

\newcommand{\naive}{\mathit{na}}

\newcommand{\semi}{\mathit{ss}}

\newcommand{\cf}{\mathit{cf}}

\newcommand{\D}{D}

\newcommand{\arrow}{\mathit{arr}}
\renewcommand{\path}{\mathit{path}}
\newcommand{\noedge}{\mathit{noe}}
\newcommand{\edge}{\mathit{e}}

\usepackage{graphicx,marvosym}

\title{Argumentative Causal Discovery}

\author{%
Fabrizio Russo\and
Anna Rapberger\and
Francesca Toni\\
\affiliations
Imperial College London, Department of Computing\\
 \emails
\{fabrizio, a.rapberger, ft\}@imperial.ac.uk
}

\begin{document}

\maketitle

\begin{abstract}
Causal discovery amounts to unearthing causal relationships amongst features in data.
It is a crucial companion to causal inference, necessary to build scientific knowledge without resorting to expensive or impossible randomised control trials.
In this paper, we explore how reasoning with symbolic representations can support causal discovery.
Specifically, we deploy \emph{assumption-based argumentation (ABA)}, a well-established and powerful knowledge representation formalism,
in combination with \emph{causality theories}, to learn graphs which reflect causal dependencies in the data. 
We prove that our method exhibits desirable properties, notably that, under natural conditions,
it can retrieve ground-truth causal graphs.
We also conduct experiments with an implementation of our method in \emph{answer set programming (ASP)} on four datasets from standard benchmarks in causal discovery, showing that our method compares well against established baselines. 
\end{abstract}

\section{Introduction}\label{intro}

Causal Discovery is the process of extracting  
causal relationships amongst variables in data, represented as graphs. 
These graphs are crucial for understanding causal effects and perform causal inference
\citep{peters2017elements,pearl2009causality,spirtes2000causation}, e.g. to determine the impact of an action or treatment on an outcome. Causal effects are ideally discovered through interventions or randomised control trials, but these can be expensive, time consuming or
outright impossible, e.g. in healthcare, trying to establish whether smoking causes cancer through a randomised control trial would require the study group to take up smoking to measure its (potentially deadly) effect. Hence the need to use observational, as opposed to interventional, data to study causes and effects \citep{peters2017elements, Schlkopf2021towards}.

Prominent approaches to perform causal discovery include constraint-based, score-based and functional causal model-based methods (see e.g. \citep{glymour2019review,vowels2022d, ZANGA2022survey} for overviews).
These approaches employ
statistical methods to retrieve the causal relations between variables. However, statistical methods, even if consistent with infinite data, are prone to errors due to finite data.
As a result, the extracted causal relations can deviate from the ground truth and, crucially, also from the observed data. 
Let us consider an example.

\begin{example}\label{ex:intro}
We set out to discover the causal relations between \emph{rain ($r$)},
\emph{wet roof terrace ($wr$)},
\emph{wet street ($ws$)} and
\emph{watering plants ($wp$)} (on the roof terrace).
After collecting sufficient data, we carry out conditional independence tests. These correctly return that $r$ and $wp$ are independent (written $r\!\indep wp$); but also find $r$ and $wp$ independent when conditioned on $\{wr\}$ (written $r\!\indep wp\!\given\!\{wr\}$) which goes against our intuition:
since something must have caused 
$wr$, we can infer $r$ when knowing $\neg wp$ and vice versa. 
That is, $r$ and $wp$ become dependent when conditioning on $\{wr\}$.

Below, we depict the ground truth causal graph (left) and the output of Majority-PC (right), proven to be sound and complete with infinite data~\citep{colombo2014order}.

\begin{center}
\begin{tikzpicture}[xscale=0.8]
                \node[cnode,minimum size=0.65cm] (1) at (0,0.5) {$r$};
                \node[cnode,minimum size=0.65cm] (0) at (0,-0.5) {$wp$};
                \node[cnode,minimum size=0.65cm] (2) at (1.1,0) {$wr$};
                \node[cnode,minimum size=0.65cm] (3) at (2.5,0) {$ws$};
5
                \path[->,thick,>=stealth]
                (1) edge (2)
                (0) edge (2)
                (2) edge (3)
                (1) edge[out=0,in=155] (3)
                ;
\begin{scope}
    [xshift=4.5cm]
                \node[cnode,minimum size=0.65cm] (1) at (0,0.5) {$r$};
                \node[cnode,minimum size=0.65cm] (0) at (0,-0.5) {$wp$};
                \node[cnode,minimum size=0.65cm] (2) at (1.1,0) {$wr$};
                \node[cnode,minimum size=0.65cm] (3) at (2.5,0) {$ws$};
                 \path[->,thick,>=stealth]
                 (1) edge[-] (2)
                 (0) edge (2)
                 (1) edge[-,out=0,in=155] (3)
                 (3) edge (2)
                 ;
\end{scope}
            \end{tikzpicture}
\end{center}

A directed edge is interpreted as cause, e.g.,  {wp} causes  {wr}; 
the absence of an edge indicates causal independence; an undirected edge indicates a causal relationship but the direction of the cause and effect relation remains unclear. 

Since the conditional independence test wrongly rendered  {r} and  {wp} independent given $\{wr\}$, it is impossible to retrieve the ground truth whilst satisfying all reported causal relations between the variables. In fact, it can happen that \emph{no} graph exists that faithfully captures the results of the tests.
\end{example}

\begin{figure*}[t]
    \centering
    \includegraphics[width=0.92\textwidth]{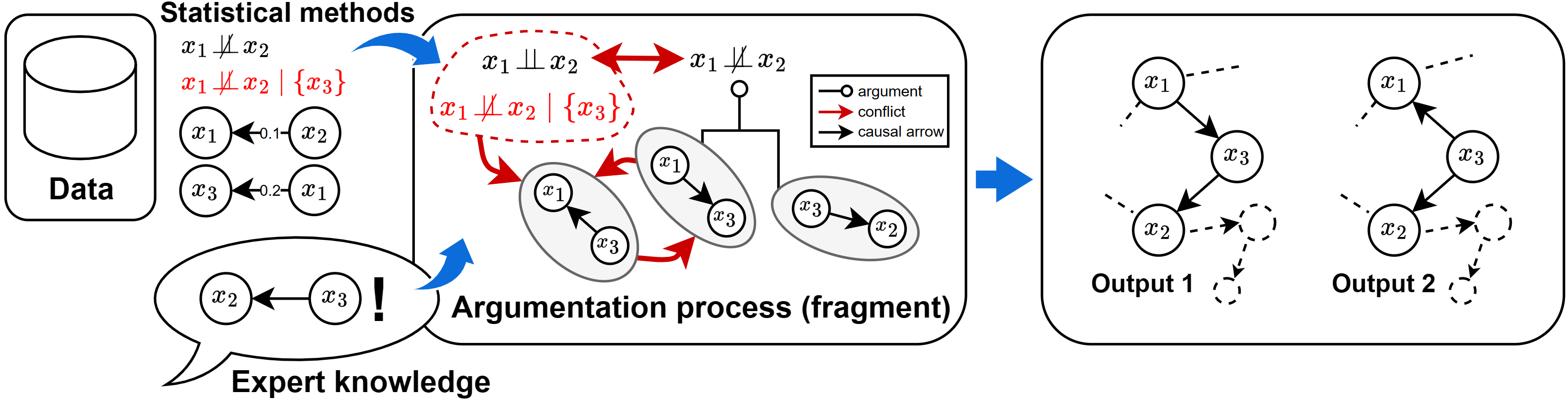}
    \caption{Overview of the workflow of our \emph{Causal ABA algorithm}, which combines statistical methods and expert domain knowledge with non-monotonic reasoning and performs argumentative reasoning to output causal graphs consistent with the reported causal relationships.}
    \label{fig:aba_workflow}
\end{figure*}
To account for the issues observed in the example, researchers have investigated several methods to handle conflicting data; e.g.,
~\cite{corander2013learning} utilised Answer Set Programming (ASP) to learn chordal Markov networks; \cite{hyttinen2014ASP} provide an encoding of graphical interventions 
to compute causal graphs; \cite{DBLP:journals/ijar/RantanenHJ20} use constraint programming.
However, argumentative methods, which are ideally suited for conflict resolution, have not received much attention in the context of causal discovery so far. 
A notable exception is the work by 
\cite{DBLP:journals/jmlr/BrombergM09} who employ a form of preference-based argumentation \citep{amgoud2002reasoning}, instantiated with deductive argumentation (see \citep{BesnardH18} for an overview), to choose a set of tests to use within the PC algorithm \citep{spirtes2000causation}.
Their method is however based on \citeauthor{pearl2009causality}'s graphoid axioms which are incomplete;
thus, some inconsistencies between the reported (in)dependences might not be detected by their approach.

In this paper, we provide a novel argumentative approach to account for inconsistencies in the reported tests and reflect a consistent subset of them into a 
\emph{directed acyclic graph (DAG)}.
In line with the causal discovery literature, we assume \emph{faithfulness} of the data, i.e., that all the independencies in the data are compatible with some DAG structure~\citep{spirtes2000causation} as well as \emph{sufficiency} i.e. there are no latent confounders.
To handle conflicts in data, we employ \emph{assumption-based argumentation (ABA)} which is a versatile non-monotonic reasoning formalism~\citep{CyrasFST2018} based on assumptions (i.e., defeasible elements) and inference rules. 
ABA has been studied under numerous 
semantics, which are criteria to determine the acceptance of assumption sets and their conclusions. A single ABA framework can possess several different extensions, i.e., sets of acceptable assumptions w.r.t.\ a given semantics, which reflect the different viewpoints that exist within a single framework.

In Fig.~\ref{fig:aba_workflow} we summarise the workflow of our method.
Based on (i) the output of statistical methods as well as potential (ii) domain knowledge provided by experts, we construct an ABA framework whose
extensions provide all the DAGs compatible with (i) and (ii). 
Overall, 
our contributions in this work are as follows:
\begin{itemize}
    \item We formalise causal graphs in the language of ABA (\emph{Causal ABA}). 
    We use rules to model the d-separation criterion~\citep{pearl2009causality}, which characterises conditional independence in DAGs; and assumptions to model conditional independence and causal relations.
\item We provide an ASP implementation of our theoretical framework using the independence tests from the Majority-PC algorithm~\citep{colombo2014order} as hard or weak constraints, resulting in \emph{ABA-PC}.
We employ weights for fact selection when necessary. 
\item We experimentally evaluate our ABA-PC algorithm with four (standard) datasets.
Our experiments show that our proposed framework improves on current state-of-the-art baselines in Causal Discovery. In particular, we reconstruct the ground-truth causal DAG better than Majority-PC using the same set of independence relations. 
\end{itemize}

 \section{Preliminaries}\label{sec:prelim}
Graphs are crucial for causal and argumentation theories.  
A graph $\DAG = (\nodeSet, \edgSet)$ has nodes $\nodeSet$ and edges $\edgSet\subseteq \nodeSet \times \nodeSet$;
$\DAG$ is \emph{directed} if either $(x,y)\in \edgSet $ or $(y,x)\in \edgSet $;
\emph{undirected} if $(x,y)\in \edgSet $
and $(y,x)\in \edgSet $; 
and \emph{partially directed} otherwise. 
The \emph{skeleton} of $\DAG$ is the result of replacing all directed edges with undirected ones.
$x,y\in \nodeSet$ are \emph{adjacent} iff $(x,y)\in \edgSet$ or $(y,x)\in \edgSet $.
A \emph{($x_1$-$x_n$-)path} $\path$ is a sequence of distinct nodes $x_1\dots x_n$ s.t.\ for $1 \leq i < n$, $x_i$ and $x_{i+1}$ are adjacent.
We omit `$x_1$-$x_n$' if it is clear from the context. 
Given a path $\Gpath=x_1\dots x_n$ and a node $x$, we sometimes abuse notation and write $x\in \Gpath$ to specify that $x$ is contained in $\Gpath$, i.e., there is $i\leq n$ s.t.\ $x=x_i$.
A path $x_1\dots x_n$ is \emph{directed} if $(x_i,x_{i+1})\in \edgSet $ for all $i\leq n$; \emph{cyclic} if it is directed and $x_1=x_n$.
A \emph{directed acyclic graph (DAG)} is a directed graph without cycles.

\subsection{Causal Graphs}
A causal graph represents causal relations between variables \citep{pearl2009causality, spirtes2000causation}.
In this paper, we focus on causal graphs that admit a DAG structure. \citeauthor{pearl2009causality}'s \emph{d-separation criterion} establishes the link between DAGs and conditional independence.

\subsubsection{Conditional (In)dependence}
We consider a finite set of \emph{variables} $\nodeSet\!$. 
For pairwise disjoint sets $\Xvar,\Yvar,\Zvar\subseteq \nodeSet$  we let  $(\Xvar\!\indep\!\Yvar\!\given\!\Zvar)$ indicate that $\Xvar$ and $\Yvar$ are \emph{independent} given the conditioning set  $\Zvar$; $(\Xvar\!\indep\!\Yvar\!\given\!\emptyset)$ is simply written as $(\Xvar\!\indep\!\Yvar)$ and singleton sets $\{x\}$ are denoted by $x$ (e.g.,
$(\{x\}\!\indep\!\{y\}\!\given\!\emptyset)$ is written as 
$(x\!\indep\!y$). 
Also,
$(\Xvar\!\notindep\!\Yvar\!\given\!\Zvar)$ means that
 $\Xvar$ and $\Yvar$ are \emph{dependent} given $\Zvar$.
A fundamental property of conditional independence is symmetry~\citep{Pearl1986GRAPHOIDSGL};
we identify $(\Xvar\!\indep\!\Yvar\!\given\!\Zvar)$ and $(\Yvar\!\indep\!\Xvar\!\given\!\Zvar)$ with each other (analogously for dependence statements). 

A triple $(x_i, x_j, x_k)$ of variables in a DAG is an \emph{Unshielded Triple (UT)} if two variables are not adjacent but each is adjacent to the third.
An UT $(x,y,z)$ is a \emph{v-structure} iff $(x,y)\in \edgSet$ and $(z,y)\in \edgSet$; 
$y$ is a \emph{collider} (w.r.t.\ $x,z$).  

\begin{definition}\label{def:active}
    Let $\DAG=(\nodeSet,\edgSet)$ be a DAG. A x-y-path $\Gpath$, $x,y\in \nodeSet$, $x\neq y$, is \emph{$\Zvar$-active} for a set $\Zvar\subseteq \nodeSet\setminus \{x,y\}$ in $\DAG$ iff 
    for each node $z\in \Gpath$: 
    if $z$ is a collider in $\Gpath$, then $z\in \Zvar$
    or there is a descendant $z'$ of $z$ s.t.\ $z'\in \Zvar$;
    otherwise, $z\notin \Zvar$.
\end{definition}
Two variables $x,y\in \nodeSet$ are \emph{independent}, conditioned on a set $\condSet \subseteq \nodeSet \setminus \{x,y\}$, if fixing the values of the variables in \condSet{} does not provide additional information about $x$ or $y$ (resp.). 
Independence in DAGs is captured by \emph{d-separation}.%
\begin{definition}
    Let $\DAG=(\nodeSet,\edgSet)$ be a DAG. Two nodes $x,y\in \nodeSet$ are \emph{d-connected} given $\Zvar\subseteq \nodeSet$ iff $\DAG$ contains a $\Zvar$-active x-y-path $\Gpath$.
    The nodes $x,y\in \nodeSet$ are \emph{d-separated} given $\Zvar$ iff $x,y$ are not d-connected given $\Zvar$.
    Two variables $x,y$ are \emph{independent} w.r.t.\ $\Zvar$ in $\DAG$ iff they are d-separated given $\Zvar$, denoted by $x\indep_{\!\DAG}\,y\given \condSet$.
\end{definition}

\subsubsection{Causal Graphs and Statistics}
Causal Discovery couples statistical and graphical methods to extract causal graphs from data.
The nodes $\nodeSet=\{X_1,\dots,X_d\}$ in a causal graph $\DAG=(\nodeSet,\edgSet)$ correspond to random variables (in our running Example~\ref{ex:intro}, `rain' can be a random variable when associated with observed data) and the edges represent causal relationships between them. 
A joint probability distribution $P$ factorizes according to a DAG $\DAG$ if $P(\nodeSet{}) = \prod_{i=1}^d P(X_i \given \text{pa}(\DAG, X_i))$, where $\text{pa}(\DAG, X_i)$ denotes the set of parents of $X_i$ in $\DAG$.  A distribution $P$ is  \emph{Markovian} w.r.t. \DAG{} if it respects the conditional independence relations entailed by $\DAG{}$ via d-separation. In turn, $P$ is \emph{faithful} to $\DAG{}$ if DAG \DAG\ reflects all conditional independences in $P$. 
Different DAGs can imply the same set of conditional independences, in which case they form a Markov Equivalence Class (MEC) \citep{Richardson1996CCD}. DAGs in a MEC present the same adjacencies and v-structures and are uniquely represented by a \emph{Completed Partially} DAG (CPDAG) \citep{chickering2002ges} which is a partially directed graph that has a directed edge if every DAG in the MEC has it, and an undirected edge if both directions appear in the MEC. 

A \emph{Conditional Independence Test (CIT)}, e.g. Fisher's Z \citep{fisher1970statistical}, HSIC \citep{Gretton2007hsic}, or KCI \citep{zhang2011kci}, is a procedure 
to measure independence with a known asymptotic distribution under the null hypothesis \nullH{} of independence. Calculating the test statistic for a dataset allows to estimate the test's observed significance level ($p$-value), under \nullH{}. This is a measure of evidence against \nullH{} \citep{hung97pvalue}. Under \altH{},  $p$  is uniformly distributed in the interval $[0,1]$, which allows to set a significance level $\alpha$ that represents the pre-experiment Type I error rate (rejecting \nullH{} when it is true), whose expected value is at most $\alpha$. 
A CIT, denoted by $I(X_i, X_j\!\given\!\condSet)$, outputs a $p$-value. If $I(X_i, X_j\!\given\!\condSet) = p \geq \alpha$ then 
$X_i\!\indep\!X_j\!\given\!\condSet$. 
Instead, if $I(X_i, X_j\!\given\!\condSet) = p < \alpha$ then we can reject \nullH{} and declare the variables dependent: $X_i\!\notindep X_j\!\given\!\condSet$.

\subsection{Assumption-based Argumentation}
We recall basics of assumption-based argumentation (ABA); for a comprehensive introduction we refer to~\citep{CyrasFST2018}.
We assume a deductive system $(\mathcal{L},\mathcal{R})$, where  $\mathcal{L}$ is a formal language, i.e., a set of sentences, 
and $\mathcal{R}$ is a set of rules over $\mathcal{L}$. A rule $r \in \mathcal{R}$ has the form
$a_0 \leftarrow a_1,\ldots,a_n$ with $a_i \in \mathcal{L}$, $head(r) = a_0$ and $body(r) = \{a_1,\ldots,a_n\}$.%
\begin{definition}
	An ABA framework (ABAF) is a tuple $(\mathcal{L},\mathcal{R},\mathcal{A},\contraryempty)$, where $(\mathcal{L},\mathcal{R})$ is a \emph{deductive system}, $\mathcal{A} \subseteq \mathcal{L}$ a set of \emph{assumptions}, and $\contraryempty:\mathcal{A}\rightarrow \mathcal{L}$ is a function mapping assumptions $a\in \mathcal{A}$ to sentences $\mathcal{L}$ (contrary function). 
\end{definition}
A sentence $q \in \mathcal{L}$ is \emph{tree-derivable} from $S \subseteq \mathcal{A}$ and rules $R \subseteq \mathcal{R}$, denoted by $S \vdash_R q$, if there is a finite rooted labeled tree $T$ which, intuitively, corresponds to the structure of the derivation:
the root of $T$ is labeled with $q$; 
the set of labels for the leaves of $T$ is equal to $S$ or $S \cup \{\top\}$; 
and for every inner node $v$ of $T$ there is a rule $r \in R$ such that $v$ is labelled with $head(r)$, the number of successors of $v$
is $|body(r)|$ and every successor of $v$ is
labelled with a distinct $a \in body(r)$ or $\top$ if $body(r)=\emptyset$.
We often drop $R$ and write $S \vdash_R q$ simply as $S \vdash q$ if it does not cause confusion.
\begin{definition}
    Let $D=(\mathcal{L},\mathcal{R},\mathcal{A},\contraryempty)$ be an ABAF.
    A set $S\subseteq \mathcal A$ \emph{attacks} $T\subseteq \mathcal A$ if there is $S'\subseteq S$, $a\in T$, s.t.\ $S'\vdash\contrary{a}$.  
A set $S$ is \emph{conflict-free} in an ABAF $D$ ($S\in\cf(D)$) if it does not attack itself;
$S$ \emph{defends} $T$ iff it attacks each attacker of $T$;
$S$ is \emph{closed} iff $S\vdash a$ implies $a\in S$; 
$S$ is \emph{admissible} ($S\in\adm(\D)$) if it is conflict-free and defends itself.
\end{definition}
With a little notational abuse we say a set $S$ of assumptions attacks an assumption $a$ if $S$ attacks the singleton $\{a\}$; we let 
$\contrary{S}=\{\contrary{a}\mid a\in S\}$.

An ABAF $D$ is called \emph{flat} iff each set $S$ of assumptions is closed. 
We call an ABAF \emph{non-flat} if it does not belong to the class of flat ABAFs.

We next recall grounded, complete, preferred, and stable ABA semantics (abbr.\ $\grd$, $\comp$, $\pref$, $\stb$).%
\begin{definition}
Let $\D$ be an ABAF and let $S\in\adm(\D)$.
\begin{itemize}
 \item $S\in \comp(\D)$ iff $S$ contains every assumption set it defends; 
 \item $S\in \grd(\D)$ iff $S$ is $\subseteq$-minimal in $\comp(\D)$;
 \item $S\in \pref(\D)$ iff $S$ is $\subseteq$-maximal in $\comp(\D)$;
 \item $S\in \stb(\D)$ iff $S$ attacks each $\{x\} \subseteq \mathcal{A} \setminus S$.
\end{itemize}
\end{definition}
Given a semantics $\sigma$, we call $\sigma(D)$ the set of \emph{$\sigma$-extensions} of the ABAF $D$. We drop `$\sigma$' if it is clear from context.

\subsubsection{Graphical ABA Representation}
\emph{Argumentation frameworks with collective attacks (SETAFs)}~\citep{nielsen2006generalization} are ideally suited to depict the attack structure between the assumptions in ABAFs as outlined by~\cite{KonigRU22}.
In brief, a SETAF is a pair $(A,R)$ consisting of a set of arguments $A$ and an attack relation $R\subseteq  2^A\times A$.
We can instantiate an ABAF 
$D=(\mathcal{L},\mathcal{A},\mathcal{R},\contraryempty)$ as SETAF by setting $A=\mathcal{A}$ and $R$ is the induced attack relation: $S\subseteq A$ attacks $a\in A$ if $S\vdash \contrary{a}$.
\begin{example}\label{ex:bg}
Consider an ABAF with assumptions $a,b,c$, their contraries $\contrary{a}=s$, $\contrary{b}=p$, $\contrary{c}=q$, and rules $(p\gets a,c)$ and $(s\gets c)$.
We can represent the ABAF as SETAF:

\begin{center}
\begin{minipage}{0.12\textwidth}
\centering
\begin{tikzpicture}[xscale=1]
\small

\path
node[] (a) at (0,0.2) {a}
node[] (b) at (1.2,0.2) {b}
node[] (c) at (0.6,-0.4) {c}
;	
\path[>=stealth,->,thick]
(a)edge[magenta,out=10,in=170,looseness=1.2](b)
(c)edge[magenta,out=110,in=170,looseness=1.6](b)
(c)edge[teal,bend left](a)
;	

\end{tikzpicture}
\end{minipage}
\hfill 
\begin{minipage}{0.34\textwidth}
\begin{itemize}
    \item $\{a,c\}$ collectively attacks $b$ as \textcolor{magenta}{$\{a,c\}\!\vdash\! p$}; 
\item
$c$ attacks $a$ since \textcolor{teal}{$\{c\}\vdash s$}.
\end{itemize}
\end{minipage}
\end{center}
The graph depicts the attack structure between the assumptions; the collective attack is depicted as a joint arrow. 
\end{example}

\section{Capturing Causal Graphs with ABA}\label{sec:ABAF narrow def}\label{sec:causal ABA}
\newcommand{\dcn}{\mathit{dc}}
\newcommand{\nocd}{\mathit{nb}}
\newcommand{\coldesc}{\mathit{cold}}
\newcommand{\nocoldesc}{\mathit{nocold}}
\newcommand{\dagg}{\mathit{dag}}
\newcommand{\dpath}{\mathit{dpath}}
\newcommand{\edgerules}{\mathit{edge}}
\newcommand{\pathrules}{\mathit{path}}
\newcommand{\cyc}{\mathit{cyc}}
We formalise causal graphs in ABA.
 We assume a fixed but arbitrary set of variables $\nodeSet$ with $|\nodeSet|=d$.
 We refrain from explicitly mentioning the language $\mathcal{L}$. Each assumption $a$ below has a distinct contrary $a_c$; for convenience, we write $\contrary{a}$ instead of $a_c$ if it does not cause confusion.

\subsection{Causal ABA}
The class of causal relations we aim to capture are characterised by two factors: acyclicity and d-separation.

\subsubsection{Acyclicity}
We formalise graph-theoretic properties since our expected outcome, i.e., the resulting extensions, are graphs. Thus, the assumptions in our ABAF are arrows:
    $$\mathcal{A}_\arrow=\{\arrow_{xy} \mid x,y\in \nodeSet,x\neq y\}.$$
Then, we define acyclicity as follows.
\begin{definition}
\label{def:acyclicity}
    Let $D_\dagg=(\mathcal{A}_\dagg, \mathcal{R}_\dagg,\contraryempty)$ where
    $$\mathcal{A}_\dagg=\mathcal{A}_\arrow\cup \{\noedge_{xy}\mid x,y\in \nodeSet,x\neq y\}$$
    and $\mathcal{R}_\dagg$ contains the following rules: 
    \begin{itemize}
    \item $\contrary{a}\gets b$, $a\neq b$, $a,b\in\{\arrow_{xy},\arrow_{yx},\noedge_{xy}\}$,
    $x,y\in \nodeSet$;
    \item $\contrary{\arrow_{x_ix_{i+1}}} \gets \arrow_{x_1x_2},\dots,\arrow_{x_{k-1}x_k}$ for each sequence $x_1\dots x_k$ with $x_1=x_k$, for each $1\leq i<k$.
    \end{itemize}
\end{definition}
Intuitively, $\noedge_{xy}$ stands for ``no edge between $x$ and $y$." Note that
we define only one atom $\noedge_{xy}$ for each pair of variables $x$, $y$.
The first set of rules enables the choice between $\noedge_{xy}$, $\arrow_{xy}$ and $\arrow_{yx}$.
The second ensures that no extension contains a cycle.

\begin{example}\label{ex:three variables example}
    Consider $\nodeSet=\{x,y,z\}$.
    The corresponding ABAF $D_\dagg$ contains 9 assumptions: for each pair of variables $u,v\in \nodeSet$, we have $\arrow_{uv}$, $\arrow_{vu}$ and $\noedge_{uv}$.
We observe that we have precisely
     two cyclic sequences of length~$>2$, namely (from $x$) $c_1=xyzx$ and $c_2=xzyx$.
    Both cycles attack each arrow it contains; the attack structure of the ABAF is depicted below.
 \begin{center}
                \begin{tikzpicture}[xscale=2,yscale=1.5]
                \node (xyar) at (0.5,-0.5) {$\arrow_{xy}$};
                \node (yxar) at (1,0) {$\arrow_{yx}$};
                \node (yxnoe) at (0,-0) {$\noedge_{xy}$};
                \node (zyar) at (2.5,-0.5)  {$\arrow_{zy}$};
                \node (yzar) at (2,0) {$\arrow_{yz}$};
                \node (yznoe) at (3,0) {$\noedge_{yz}$};
                \node (xzar) at (1.5,-1) {$\arrow_{xz}$};
                \node (zxar) at (1,-1.5) {$\arrow_{zx}$};
                \node (zxnoe) at (2,-1.5) {$\noedge_{xz}$};
                \path[<->,>=stealth]
                (xyar) edge (yxar)
                (xyar) edge (yxnoe)
                (yxnoe) edge (yxar)
                (zyar) edge (yzar)
                (zyar) edge (yznoe)
                (yznoe) edge (yzar)
                (xzar) edge (zxar)
                (xzar) edge (zxnoe)
                (zxnoe) edge (zxar)
                ;
                \path[->,cyan,>=stealth]
                (xyar) edge[out=-90,in=0,looseness=2.6] (xyar)
                (yzar) edge[out=190,in=0] (xyar)
                (zxar) edge[out=90,in=0] (xyar)
                ;
                \path[->,>=stealth,green!80!black]
                (xyar) edge[out=20,in=175] (yzar)
                (yzar) edge[out=280,in=175,looseness=2.6] (yzar)
                (zxar) edge[out=70,in=175] (yzar)
                ;
                \path[->,blue!70!cyan,>=stealth]
                (xyar) edge[out=-60,in=110] (zxar)
                (yzar) edge[out=200,in=110] (zxar)
                (zxar) edge[out=180,in=110,looseness=2.2] (zxar)
                ;

                \path[->,magenta,>=stealth]
                (yxar) edge[out=200,in=-70,looseness=2.8] (yxar)
                (xzar) edge[out=150,in=-70] (yxar)
                (zyar) edge[out=180,in=-70] (yxar)
                ;
                 \path[->,orange!70!black,>=stealth]
                (yxar) edge[out=-20,in=-140,looseness=0.7] (zyar)
                (xzar) edge[out=-10,in=-140] (zyar)
                (zyar) edge[out=-70,in=-140,looseness=4.5] (zyar)
                ;
                \path[->,>=stealth,purple!80!blue,thick]
                (yxar) edge[out=0,in=70] (xzar)
                (xzar) edge[out=10,in=70,looseness=3.4] (xzar)
                (zyar) edge[out=160,in=70] (xzar)
                ;
                
            \end{tikzpicture}
\end{center}  

The joint arcs represent collective attacks; e.g., the thick, purple arrows pointing to $\arrow_{xz}$ represent the attack from set $\{\arrow_{yx},\arrow_{xz},\arrow_{zy}\}$ on the assumption $\arrow_{xz}$ based on the derivation $\{\arrow_{yx},\arrow_{xz},\arrow_{zy}\}\vdash \contrary{\arrow_{xz}}$.
\end{example}

We show that $D_\dagg$ correctly captures the set of all DAGs of fixed size $d$. The correspondence is true for all (except $\grd$) argumentation semantics under consideration. Below, we use the assumption $\arrow_{xy}$ to stand for the arrow $(x,y)$.
All proofs of this section are provided in Appendix \S\ref{sec:proof}.
\begin{proposition}[restate=PropDAGABA,name=]
    $\{(\nodeSet,S\cap \mathcal{A}_\arrow)\mid S\in \sigma(D_\dagg)\}=\{\DAG\mid \DAG \text { is a DAG}\}$ for $\sigma\in\{\com,\prf,\stb\}$.
\end{proposition}
Note that the grounded extension corresponds to the fully disconnected graph $\DAG=(\nodeSet,\emptyset)$ since the empty set is complete. 
Note also that the correspondence between DAGs and the extensions of the ABAF is one-to-many for complete, admissible and conflict-free assumption sets since a single acyclic graph corresponds to several complete extensions. 
Accepting the absence of an edge between two variables $x,y$ can be realised by accepting $\noedge_{xy}$ or simply by accepting none of $\noedge_{xy}$, $\arrow_{xy}$, $\arrow_{yx}$ in the extension. 
\begin{example}\label{ex:extensions for the fully disc graph}
    In the ABAF from Example~\ref{ex:three variables example}, the fully disconnected graph $(\nodeSet,\emptyset)$ corresponds to $2^3$ complete extensions; i.e, to each subset of $\{\noedge_{xy}, \noedge_{yz},\noedge_{zx}\}$. 
\end{example}
For preferred and stable semantics, the correspondence is one-to-one; the semantics coincide in $D_\dagg$, as stated below.%
\begin{lemma}[restate=LemCausalABAStablecoincide,name=]
\label{lem:ABAcausal stable pref etc coincide}
 $\sigma(D_\dagg)=\tau(D_\dagg)$ for $\sigma,\tau\!\in\!\{\prf,\stb\}$.
\end{lemma}

\begin{corollary}
Let $\sigma\!\in\!\{\prf,\stb\}$. Each DAG~$\DAG$ corresponds to a unique set $S\!\in\! \sigma(D_\dagg)$ and vice versa.
\end{corollary}

\subsubsection{D-separation}
\newcommand{\final}{V}
\newcommand{\ind}{\mathit{ind}}
\newcommand{\dsep}{\mathit{ds}}
\newcommand{\activepath}{\mathit{ap}}
\newcommand{\blockedpath}{\mathit{bp}}

The first step to represent d-separation is to extend our ABAF with independence statements. 
We do so by assuming independence between variables. We let 
$$\mathcal{A}_\ind= \{(x\!\indep\!y\!\given\!\Zvar)\mid \Zvar\subseteq \nodeSet, x,y\in \nodeSet\setminus \Zvar,x\neq y\}.$$
The conditional independence $x\!\indep\!y\!\given\!\Zvar$ is violated if the variables $x$, $y$ are d-connected, given the conditioning set $\Zvar$.
Intuitively, we want to formalise
$$\contrary{x\!\indep\!y\!\given\!\Zvar}\text{ if there exists a }\Zvar\text{-active path between }x,y.$$ 
\newcommand{\graph}{\mathit{graph}}
To capture this, it is convenient to formalise directed paths and we do so by letting $\mathcal{R}_\graph$  contain the following rules:
\begin{align*}
    \dpath_{xy}\gets  \arrow_{xy} \qquad 
    \dpath_{xz}\gets  \dpath_{xy},\arrow_{yz}\\
    \edge_{xy}\gets \arrow_{xy} \qquad \edge_{xy}\gets \arrow_{yx} \qquad \contrary{\noedge_{xy}}\gets \edge_{xy}
\end{align*}

where, intuitively, $\edge_{xy}$ stands for  ``edge between $x$ and $y$."

To formalise d-connectedness in the context of ABA, we {introduce} \emph{collider-trees}, which generalise the {notion of path} by adding branches from collider nodes.  

\begin{definition}
Let $\DAG=(\nodeSet,\edgSet)$ be a DAG, $x,y\in \nodeSet$.
    A \emph{x-y-collider-tree} $\tree$ is a sub-graph of $\DAG$ satisfying:
    \begin{itemize}
        \item[(a)] $\tree$ contains an x-y-path $\ctpath$;
        \item[(b)] for all $u\in \tree$, if $u\notin \ctpath$ then there is  $v\in \tree$ such that $v$ is a collider in $\ctpath$ and $u$ is a descendant of $v$.
    \end{itemize}
A c-t-path from collider node $c$ (of $\ctpath$) to a leaf node $t$, $t\notin \{x,y\}$, is called a \emph{branch} of $\tree$.
 For a set of variables $\Zvar\subseteq \nodeSet$, we call
    $\tree$
     \emph{\Zvar-active} iff $\ctpath$ is active w.r.t.\ $\tree\cup (\nodeSet,\emptyset)$.
\end{definition}
In the remainder of the paper, we drop `x-y' and simply say collider-tree whenever it does not cause confusion. 

\begin{example}
Consider a causal graph $\DAG$ with five variables $x,y,z,u,v$ as depicted below (left), and some collider-trees, {denoted} $\Gpath_1$, $\Gpath_2$, $\Gpath_3$, {from top to bottom, }resp.:
\begin{center}
\begin{tikzpicture}[xscale=0.8]
                \node[cnode] (x) at (0,0.4) {$x$};
                \node[cnode] (y) at (0,-0.4) {$y$};
                \node[cnode] (z) at (1.1,0) {$z$};
                \node[cnode] (u) at (2.5,0) {$u$};
                \node[cnode] (v) at (3.9,0) {$v$};

                \path[->,thick,>=stealth]
                (y) edge (x)
                (y) edge (z)
                (x) edge (z)
                (z) edge (u)
                (u) edge (v)
                ;
\begin{scope}
    [xshift=5cm,yshift=0.2cm]
 
    \node (x) at (0,0) {$x$};
    \node (u) at (4.2,0) {$u$};
    \node (z) at (2.8,0) {$z$};
    \node (y) at (1.4,0) {$y$};

    \path[->,thick,>=stealth]
    (y) edge (x)
    (z) edge (u)
    (y) edge (z)
    ;
    
    \begin{scope}
        [yshift=0.35cm]
        \node (x) at (0,0) {$x$};
        \node (y) at (1.4,0) {$y$};

        \path[->,thick,>=stealth]
        (y) edge (x)
        ;
    \end{scope}
    
    \begin{scope}
        [yshift=-0.35cm]
        \node (x) at (0,0) {$x$};
        \node (y) at (2.8,0) {$y$};
        \node (z) at (1.4,0) {$z$};
        \node (u) at (2.8,-.4) {$u$};
        \node (v) at (4.2,-.4) {$v$};

        \path[->,thick,>=stealth]
        (y) edge (z)
        (x) edge (z)
        (z) edge[in=175,out=-35] (u) 
        (u) edge (v)
        ;
    \end{scope}
    
\end{scope}
\end{tikzpicture}     
\end{center}

The collider-trees $\Gpath_1$ and $\Gpath_2$ are active w.r.t.\ $\emptyset$; both paths have no collider so they are active w.r.t.\ every set not intersecting them; $\Gpath_3$ is active for sets containing $z$, $u$ or $v$.
\end{example}
We are now ready to define our causal ABA framework. 
\newcommand{\actv}{\mathit{act}}
\begin{definition}\label{def:Dpearl}
    A \emph{causal ABAF} $D_\dsep=(\mathcal{A}_\dsep,\mathcal{R}_\dsep,\contraryempty)$ is characterised by $$\mathcal{A}_\dsep= \mathcal{A}_\dagg\cup\mathcal{A}_\ind, \text{ and } \mathcal{R}_\dsep=\mathcal{R}_\dagg\cup \mathcal{R}_\graph\cup\mathcal{R}_\actv,$$ where $\mathcal{R}_\actv$ contains  rules $(\contrary{x\!\indep\!y\!\given\!\Zvar}\gets \tree)$
    for each $\Zvar$-active x-y-collider-tree $\tree$ with $x\neq y$,
and $\Zvar\subseteq \nodeSet\setminus \{x,y\}$.
\end{definition}

\begin{example}\label{ex:ABAF explanation example}
Let us consider again Example~\ref{ex:three variables example} with $\nodeSet=\{x,y,z\}$. 
We extend our ABAF with six independence assumptions and add the corresponding contraries. Below, we depict all arguments and attacks for the pair $x,y$; i.e., all attacks on the new assumptions $(x\!\indep\!y)$ and $(x\!\indep\!y\!\given\!\{z\})$. 

 \begin{center}
\begin{tikzpicture}[xscale=2,yscale=1.5]
                \node (xy) at (0.7,1) {$x\!\indep\!y$};
                \node (xyz) at (2.3,1) {$x\!\indep\!y\!\given\!\{z\}$};
                \node (xyar) at (0.5,-0.5) {$\arrow_{xy}$};
                \node (yxar) at (1,0) {$\arrow_{yx}$};
                \node (yxnoe) at (0,-0) {$\noedge_{xy}$};
                \node (zyar) at (2.5,-0.5)  {$\arrow_{zy}$};
                \node (yzar) at (2,0) {$\arrow_{yz}$};
                \node (yznoe) at (3,0) {$\noedge_{yz}$};
                \node (xzar) at (1.5,-1) {$\arrow_{xz}$};
                \node (zxar) at (1,-1.5) {$\arrow_{zx}$};
                \node (zxnoe) at (2,-1.5) {$\noedge_{xz}$};
                
                \path[<->,>=stealth,black!30]
                (xyar) edge (yxar)
                (xyar) edge (yxnoe)
                (yxnoe) edge (yxar)
                (zyar) edge (yzar)
                (zyar) edge (yznoe)
                (yznoe) edge (yzar)
                (xzar) edge (zxar)
                (xzar) edge (zxnoe)
                (zxnoe) edge (zxar)
                ;
                \path[->,>=stealth,black!30]
                (xyar) edge[out=-90,in=0,looseness=2.6] (xyar)
                (yzar) edge[out=190,in=0] (xyar)
                (zxar) edge[out=90,in=0] (xyar)
                (xyar) edge[out=20,in=175] (yzar)
                (yzar) edge[out=280,in=175,looseness=2.6] (yzar)
                (zxar) edge[out=70,in=175] (yzar)
                (xyar) edge[out=-60,in=110] (zxar)
                (yzar) edge[out=200,in=110] (zxar)
                (zxar) edge[out=180,in=110,looseness=2.2] (zxar)
                (yxar) edge[out=200,in=-70,looseness=2.8] (yxar)
                (xzar) edge[out=150,in=-70] (yxar)
                (zyar) edge[out=180,in=-70] (yxar)
                (yxar) edge[out=-20,in=-140,looseness=0.7] (zyar)
                (xzar) edge[out=-10,in=-140] (zyar)
                (zyar) edge[out=-70,in=-140,looseness=4.5] (zyar)
                (yxar) edge[out=0,in=70] (xzar)
                (xzar) edge[out=10,in=70,looseness=3.4] (xzar)
                (zyar) edge[out=160,in=70] (xzar)
                ;

                \path[->,>=stealth,cyan,thick]
                (xyar) edge[out=100,in=-150] (xy)
                (yxar) edge[out=140,in=-130] (xy)
                (xzar) edge[out=100,in=-90] (xy)
                (zyar) edge[out=170,in=-90] (xy)
                (yzar) edge[out=190,in=-50] (xy)
                (zxar) edge[out=100,in=-50] (xy)
                (zyar) edge[out=190,in=-30] (xy)
                (zxar) edge[out=80,in=-30] (xy)
                ;
                \path[->,>=stealth,magenta,thick]
                (xyar) edge[out=0,in=-90] (xyz)
                (yxar) edge[out=10,in=-130] (xyz)
                (xzar) edge[out=0,in=-60] (xyz)
                (yzar) edge[out=20,in=-60] (xyz)
                ;
                
            \end{tikzpicture}
\end{center} 
The assumption $(x\!\indep\!y)$ is attacked by $\arrow_{xy}$, $\arrow_{yx}$ and by all x-y-paths with inner node $z$ except for the collider; $(x\!\indep\!y\!\given\!\{z\})$ is attacked by $\arrow_{xy}$, $\arrow_{yx}$ and the collider  $\{\arrow_{xz},\arrow_{yz}\}$. 
Attacks for the other pairs are analogous. 
\end{example}

The formalization correctly captures independence, as stated in the following proposition.

\begin{proposition}[restate= PropCausalABACorrectness,name=]
\label{prop:ABA Dsep semantics correspondence}
Let $\sigma\!\in\!\{\prf,\stb\}$, $S\in\sigma(D_\dsep)$, $x,y\in \nodeSet$, $\Zvar\subseteq \nodeSet\setminus \{x,y\}$. Then $(x\!\indep\!y\!\given\!\Zvar)\in S$ iff $(x\indep\!_\DAG\,y\given \Zvar)$. 
\end{proposition}

Note that we cannot guarantee the correspondence for complete semantics, as illustrated next. 

\begin{example}
    In the ABAF from Example~\ref{ex:ABAF explanation example}, $S=\emptyset$ is complete; indeed, $D_\dsep$ does not contain assumptions that are unattacked.
    The corresponding graph is $\DAG=(\nodeSet,\emptyset)$ (cf.\ Example~\ref{ex:extensions for the fully disc graph}). In $\DAG$, each pair of variables is independent; however, $S$ does not contain any independence statement.
\end{example}
The example above shows that the correspondence between independence assumptions and independencies entailed by a DAG via d-separation is not preserved when dropping $\subseteq$-maximality of the extensions. 
Interestingly, the other direction of Proposition~\ref{prop:ABA Dsep semantics correspondence} still holds for complete semantics; i.e., 
no incorrect independence statements are included in a complete extension.

\begin{proposition}[restate=PropCausalABAforAdmComCf, name=]
Let $S\in\com(D_\dsep)$, $x,y\in \nodeSet$, $\Zvar\subseteq \nodeSet\setminus \{x,y\}$. Then $(x\!\indep\!y\!\given\!\Zvar)\in S$ implies $(x\!\indep\!_\DAG\,y\!\given\!\Zvar)$. 
\end{proposition}

\subsection{
Integrating Causal Knowledge}

So far, we have introduced an ABAF that faithfully captures conditional independence in causal models.
We have shown that an independence statement $(x\!\indep\!y\!\given\!\Zvar)$ is contained in an extension $S$ if and only if it is consistent with graph corresponding to $S$. 
This, in turn translates to the dependencies of the graph:  $x$ and $y$ are dependent given $\Zvar$ iff $(x\!\indep\!y\!\given\!\Zvar)\notin S$. 

Our proposed ABAF can be integrated in any causal discovery pipeline to add formal guarantees that the graph discovered corresponds to the independences in the data.
We integrate information from external sources, might they be statistical methods or  experts, as \emph{facts}.\footnote{Assuming sufficient accuracy of the data as a first step; later on, we will assign weights to the reported independence statements in our final system to account for statistical errors.}

In the remainder of this section, we write $D\cup \{r\}=(\mathcal{A},\mathcal{R}\cup \{r\},\contraryempty)$ for an ABAF $D=(\mathcal{A},\mathcal{R},\contraryempty)$ and rule $r$.

Let us consider again our three-variables example from before (cf.\ Example~\ref{ex:ABAF explanation example}). First, suppose we have learned that $x$ and $y$ are marginally independent. We incorporate this information simply by adding the rule 
$(x\!\indep\!y\gets).$
This rule ensures that each extension contains $(x\!\indep\!y)$. Since each extension must be closed, no active path between $x$ and $y$ can be accepted. 
We can proceed similarly when incorporating specific causal relations (directed edges).
\begin{proposition}[restate=PropIntegrateDirectedEdges,name=]\label{prop:causal ABA add the remaining facts}
    Let $x,y,a,b\in \nodeSet$, $\Zvar\subseteq \nodeSet\setminus \{x,y\}$, $\Xvar\subseteq \nodeSet\setminus\{a,b\}$, $\sigma\!\in\!\{\prf,\naive,\semi,\stage,\stb\}$, $r\in \{(x\!\indep\!y\!\given\!\Zvar \gets),$ $(\arrow_{xy}\gets)\}$. For each $S\in \sigma(D_\dsep\cup \{r\})$, it holds that $$(a\!\indep\!b\!\given\!\Xvar)\in S\text{ iff }(a\!\indep\!_\DAG\,b\given\!\Xvar).$$
\end{proposition}

Crucially, we observe that adding external facts comes at a cost: the ABAF is not flat anymore; indeed, the independence and arrow literals might appear in the head of rules. 

Now, what happens if we incorporate test results or causal relation from an external source? 
Suppose we discovered $x$ and $y$ are marginally dependent. When we add the rule $(\contrary{x\!\indep\!y}\gets)$ we successfully render $(x\!\indep\!y)$ false; however, we lose the correspondence between the (in)dependence statements and the graph of a given extension: when adding the contrary of $(x\!\indep\!y)$ nothing (in the framework presented so far) prevents us from accepting one of the arrows $\arrow_{xy}$ or $\arrow_{yx}$.
We need to generalise the framework to ensure that our ABAF is sound when adding dependencies as facts to the framework, as discussed next. 

\newcommand{\dsepbp}{\mathit{csl}}
\newcommand{\dsepxyZ}{\mathit{ext}}
\newcommand{\xyZ}{\mathit{xyZ}}
\subsubsection{Blocked paths}
The ABAF $D_\dsep$ successfully captures that an active path implies dependence. 
To guarantee soundness, it remains to formalise the other direction: independence between two nodes $x,y$ implies that each path linking them is blocked. For this, we introduce new assumptions $$\mathcal{A}_\blockedpath=\{\blockedpath_{\Gpath\mid\Zvar}\mid \Gpath\text{ is a x-y-path},\Zvar\subseteq \nodeSet\setminus\{x,y\}\}$$
with contraries 
$\contrary{\blockedpath_{\Gpath\mid\Zvar}}=\activepath_{\Gpath\mid\Zvar}.$

Furthermore, we require two new sets of rules: the first set of rules formalises that the independence between two variables $x$ and $y$ given $\Zvar$ requires that each path between $x,y$ is blocked; the second set specifies when a path is $\Zvar$-active. 
\begin{definition}\label{def:causal ABAF: more rules}
        For $x,y\in \nodeSet$, $x\neq y$, $\Zvar\subseteq \nodeSet\setminus\{x,y\}$, we define $\mathcal{R}_{\dsepxyZ}=\mathcal{R}_\dsep\cup \mathcal{R}_{\xyZ}$ with $\mathcal{R}_{\xyZ}$ containing the rules 
    \begin{itemize}
        \item $x\!\indep\!y\!\given\!\Zvar \gets \blockedpath_{\Gpath_1\mid \Zvar},\dots,\blockedpath_{\Gpath_k\mid \Zvar}$ where $\Gpath_1,\dots,\Gpath_k$ denote all paths between $x$ and $y$;
        \item $\activepath_{\Gpath\mid\Zvar}\gets \ctpath$ for each $\Zvar$-active x-y-collider-tree $\tree$ with underlying x-y-path $\ctpath$. 
    \end{itemize}
\end{definition}
Let us consider the effect of these rules with an example.
\begin{example}
Consider again Example~\ref{ex:ABAF explanation example}; 
suppose we observed $x\!\notindep\!y\!\given\!\{z\}$. We add  the independence $(\contrary{x\!\indep\!y\!\given\!\{z\}}\gets)$ which prevents us from accepting all $\blockedpath_{\Gpath\mid \{z\}}$ assumptions at the same time (since each extension $S$ is closed, we also accept $(x\!\indep\!y\!\given\!\{z\})$, therefore, this leads to a conflict).  Consequently, one of the $\blockedpath_{\Gpath\mid \{z\}}$ assumptions is attacked, i.e., some path between $x,y$ is active.

It can be checked that the paths $\Gpath_1$, $\Gpath_2$, $\Gpath_3$ depicted below are $\{z\}$-active:\begin{center}
\begin{tikzpicture}[xscale=0.7]
                \node (x) at (0,0) {$x$};
                \node (y) at (1.4,0) {$y$};

                \path[->,thick,>=stealth]
                (x) edge (y)
                ;
                \begin{scope}
                    [xshift=3cm]
                \node (x) at (0,0) {$x$};
                \node (y) at (1.4,0) {$y$};

                \path[->,thick,>=stealth]
                (y) edge (x)
                ;
                \end{scope}
                
                \begin{scope}
                    [xshift=5.9cm]
                \node (x) at (0,0) {$x$};
                \node (y) at (2.8,0) {$y$};
                \node (z) at (1.4,0) {$z$};

                \path[->,thick,>=stealth]
                (x) edge (z)
                (y) edge (z)
                ;
                \end{scope}
\end{tikzpicture}  
\end{center}
As visualised in Example~\ref{ex:ABAF explanation example}, each of these paths attack $(x\!\indep\!y\!\given\!\{z\})$. Due to the new rules from Definition~\ref{def:causal ABAF: more rules} each path $\Gpath_i$ also derives $\activepath_{\Gpath_i\mid\Zvar}$ which attacks $\blockedpath_{\Gpath_i\mid\Zvar}$. Therefore, each extension $S$ must contain one of these paths.
\end{example}

We note that it suffices to add rules only for the dependence fact that we want to add.
That is, when introducing fact  $(\contrary{x\!\indep\!y\!\given\!\Zvar}\gets)$ it suffices to add the rules from Definition~\ref{def:causal ABAF: more rules} for $x$, $y$, $\Zvar$. 
We define the extended ABAF.
\begin{definition}
    \label{def:causal ABA: Dpearl refined}
    For $x,y\in \nodeSet$, $\Zvar\subseteq \nodeSet\setminus\{x,y\}$, the extended causal ABAF $D_\dsepbp^{xy\Zvar}=(\mathcal{A}_{\dsepbp},\mathcal{R}_\dsepbp,\contraryempty)$ is characterised by $\mathcal{A}_\dsepbp= \mathcal{A}_\dsep\cup\mathcal{A}_\blockedpath$ and $\mathcal{R}_\dsepbp= \mathcal{R}_\dsepxyZ\cup \{\contrary{x\!\indep\!y\!\given\!\Zvar}\gets \}$.
\end{definition}
The ABAF is sound and complete, as stated below.
\begin{proposition}[restate=PropCABAsinglestatement,name=]\label{prop:causal ABA add only facts}
Let $x,y,a,b\in \nodeSet$, 
let $\Zvar\subseteq \nodeSet\setminus\{x,y\}$ and
$\Xvar\subseteq \nodeSet\setminus\{a,b\}$,
let $\sigma\!\in\!\{\prf,\stb\}$ and let
    $S\in \sigma(D_\dsepbp^{xy\Zvar})$.
    It holds that $(a\!\indep\!b\!\given\!\Xvar)\in S\text{ iff }(a\!\indep\!_\DAG\,b\!\given\!\Xvar).$
\end{proposition}
Together, Propositions~\ref{prop:causal ABA add the remaining facts} and~\ref{prop:causal ABA add only facts} guarantee that causal knowledge can be integrated in a faithful way.
We obtain that this fine-tuned specification allows us to add (in)dependence facts and arrows whilst guaranteeing consistency of the causal ABAF.
Independence facts and arrows can be added without further changes to the framework; when adding dependence facts, we require additional rules as specified in Definition~\ref{def:causal ABAF: more rules}.
Below, we denote by $D_\dsepbp^T$, where $T$ is a set of (in)dependence and arrow facts, 
the ABAF obtained by the iterative update of the ABAF $D_\dsep$ with $D_\dsepbp^{xy\Zvar}$ for all dependence facts $(\contrary{x\!\indep\!y\!\given\!\Zvar}\gets)\in T$.
\begin{corollary}
Let $T$ be a set of (in)dependence and arrow facts, $\sigma\!\in\!\{\prf,\stb\}$, $x,y\in \nodeSet$, $\Zvar\subseteq \nodeSet\setminus\{x,y\}$ and $S\in \sigma(D_\dsepbp^T)$.
    Then $(x\!\indep\!y\!\given\!\Zvar)\in S\text{ iff }(x\indep\!_\DAG\, y\given \Zvar).$
\end{corollary}

\section{Implementation}
In this section, we present an instance of our \emph{Causal ABA algorithm} which combines our causal ABAF with heuristic approaches to select the independence facts that it can take in input. 
The workflow of Algorithm~\ref{alg:ABAPC} is as follows:
\begin{enumerate}
    \item The main function of the algorithm is what we name $\texttt{\textbf{causalaba}}$ (line 9 and 12 of Algorithm~\ref{alg:ABAPC}). The causal ABAF instance is determined by the number of nodes in the graph $d$ and a set of facts $\mathbf{T}$. We \emph{generate the causal ABAF} $D_\dsepbp^\mathbf{T}$ presented in \S\ref{sec:ABAF narrow def}, using 
    an ASP implementation in clingo~\citep{GebserKKS17}.
    We then compute the stable extensions of the causal ABAF. Our ASP encoding is detailed in \S\ref{subsec:ASP}. 
    \item The main input of Algorithm~\ref{alg:ABAPC} is a set of independence facts ($\mathcal{I}$), alongside the significance threshold $\alpha$ and the number of nodes $d$. We discuss \emph{sourcing facts} in \S\ref{subsec:sourcing facts}.
    \item  As shown in \S\ref{sec:causal ABA}, each stable extension corresponds to a DAG compatible with the fixed set of independence tests. However, statistical methods can return erroneous results, in which case our causal ABAF might output no stable extension at all.
    To overcome this problem, we \emph{select facts} by assigning them appropriate weights (lines 2-12) and use these weights both to optimise (using weak constraints within \texttt{\textbf{causalaba}}) and rank (possibly several) output extensions (lines 10-20). We discuss this in \S\ref{subsec:fact selection}.
\end{enumerate}
  
\begin{algorithm}[t]
    \caption{Causal ABA (with independence facts)}
    \label{alg:ABAPC}
    \textbf{Input}: $\mathcal{I}, \alpha, |\nodeSet|=d$ \\[-1em]
        \begin{algorithmic}[1]
            \State $\mathbf{T} \leftarrow 
            [\hspace{0.15cm}]$
            \For{$p=I(x,y\given\condSet) \in \mathcal{I}$
            }
                \State $s\leftarrow |\condSet|$
                \If{$p > \alpha$}
                \State \hspace{-0.4cm}$\mathbf{T}\leftarrow\mathbf{T}+[(\texttt{\textbf{indep}}(x,y,\condSet), \mathcal{S}(p, \alpha, s,d))]$ 
                \Else 
                \State \hspace{-0.4cm}$\mathbf{T} \leftarrow \mathbf{T}+[(\texttt{\textbf{dep}}(x,y,\condSet), \mathcal{S}(p, \alpha, s, d))]$
                \EndIf
            \EndFor
            \State $\mathbf{T} \leftarrow \text{sort}(\mathbf{T}, \mathcal{S})$\!\!\!\Comment{Sort elements of $\mathbf{T}$ by strength $\mathcal{S}$}
            \State $\mathbf{M} = \texttt{\textbf{causalaba}}(d, \mathbf{T})$
            \While{$\mathbf{M}=\emptyset$}
               \State $\mathbf{T} \leftarrow \mathbf{T}[2\ldots|\mathbf{T}|]$ \Comment{Drop fact with lowest $\mathcal{S}$}
               \State $\mathbf{M} = \texttt{\textbf{causalaba}}(d, \mathbf{T})$
            \EndWhile
            \For{$\DAG \in \mathbf{M}$}
                \State $\mathcal{S}_\DAG = 0$
                \For{$p=I(x,y\given\condSet) \in \mathcal{I}$}
                    \State $s\leftarrow |\condSet|$
                    \If{$p > \alpha \And x\indep_{\!\DAG}\, y\given \condSet$} 
                        \State $\mathcal{S}_\DAG \leftarrow \mathcal{S}_\DAG + \mathcal{S}(p, \alpha, s,d)$
                    \Else
                        \State $\mathcal{S}_\DAG \leftarrow \mathcal{S}_\DAG - \mathcal{S}(p, \alpha, s,d)$
                    \EndIf
                \EndFor
            \EndFor
            \State $\DAG \leftarrow \text{argmax}(\DAG \in \mathbf{M}, \mathcal{S}_\DAG)$\!\! \Comment{Select $\DAG$ with max $\mathcal{S}_\DAG$}
            \Return $\DAG$
        \end{algorithmic}
\end{algorithm}

Our proposed Algorithm~\ref{alg:ABAPC} is a sound procedure to extract DAGs given a consistent set of independencies.
\begin{proposition}
Given a set $\nodeSet$ of variables and a set of (in)dependencies $\mathcal{I}$, compatible with a (set of) MEC(s),
Algorithm~\ref{alg:ABAPC} outputs a DAG consistent with $\mathcal{I}$.
\end{proposition}

In this work, we instantiate our Algorithm~\ref{alg:ABAPC} using the Majority-PC algorithm (MPC)~\citep{colombo2014order} to source facts, resulting in the \emph{ABA-PC algorithm}.
In the following subsections, we detail our implementation.

\begin{remark}
    The causal ABAF $D_\dsepbp^T$ from Definition~\ref{def:causal ABA: Dpearl refined} is potentially \emph{non-flat} since assumptions can be derived:  independence assumptions as well as $\arrow$ and $\activepath$ assumptions may appear in the head of rules.
    Thus, it lies in a broader ABA class, affecting semantical properties known for flat ABAFs; for instance, complete extensions may not always exist~\citep{CyrasFST2018,DBLP:conf/aaai/0001PRT24}. As a consequence, standard ABA solvers are not applicable to our case since they typically focus on the class of flat ABAFs.
    In this work, we therefore propose an Answer Set Programming (ASP) encoding 
    of our causal ABAF under stable semantics. 
    This also allows us to exploit ASP's grounding abilities to obtain causal ABAFs from concise schemata representations (see~\citep{ABALearn} for the presentation of ABA in terms of schemata).
\end{remark}

\subsection{Encoding Causal ABA in ASP}\label{subsec:ASP}

Stable ABA semantics and stable semantics for Logic Programs (LP) are closely related~\citep{CaminadaSAD15assum,DBLP:conf/aaai/0001T15}; crucially, their correspondence has recently been extended to non-flat instances~\citep{RapbergerUT24}.
In standard ABA-LP translations, assumptions are associated with their default negated contraries: an assumption $a\in\mathcal{A}$ with contrary $a_c\in\mathcal{L}$ corresponds to the default negated literal $\textit{not}\ a_c$. 
These translations, however, consider only the case where the underlying logical language is atomic.
To exploit the full power of ASP, we slightly deviate from standard translations, when appropriate, whilst guaranteeing consistency with our model.
We also make use of more descriptive contrary names to enable a more intuitive reading.

For a set of variables $\nodeSet$, we express the causal ABAF by 
\begin{enumerate}
    \item encoding DAGs: each answer set corresponds to a DAG; 
    \item encoding d-separation: nodes $x$ and $y$ are independent given $\Zvar$ \emph{iff} $x$ and $y$ are not linked via an active path. 
\end{enumerate}
Following the standard translation, each ABA atom $\arrow_{xy}$ is translated to $\textit{not}\ \contrary{\arrow_{xy}}$.
Here, we identify ``not $\arrow_{yx}$'' simply with \textbf{arrow}(x,y) and ``not $\noedge_{xy}$'' with \textbf{edge}(x,y).
In our encoding, each answer set corresponds to precisely one DAG of size $|\nodeSet|=d$ for a given set $\nodeSet$ of variables. We encode acyclicity and further DAG-specific elements as expected;
the encoding is given in the Appendix \S\ref{sec:asp_det}.

To link causality and DAGs we encode the d-separation criterion.
To handle sets in ASP, we encode the ($k$-th) set $S\subseteq \nodeSet$ with predicates \textbf{in}(k,x). 
Module $\aspmodule{col}$ in Listing~\ref{asp:col} encodes collider and collider descendant (with natural specifications of the \textbf{arrow} and \textbf{dpath} (directed path) predicates).
Next, we introduce \emph{non-blocking} nodes: node $v\notin \{x,y\}$ in an x-y-path  is non-blocking, given $\Zvar$, iff 
\begin{itemize} 
    \item 
    $v$ is a collider (with respect to its neighbours in the path) and either $v\in \Zvar$ or a descendant of $v$ is in $\Zvar$; or
    \item $v$ is not a collider and $v\notin \Zvar$.
\end{itemize}
Lines 3-5 in 
Module $\aspmodule{col}$ in Listing~\ref{asp:nonblocker} encode these rules. 
Now, for each pair $x,y\in \nodeSet$, for each set $S\setminus \{x,y\}$, for each x-y-path $\Gpath=v_1\dots v_n$ with $x=v_1$ and $y=v_n$, we add rules $\Pi_{ap}(\Gpath,(v_i)_{i\leq k})$ as specified in Listing~\ref{asp:activepath}.
\begin{listing}[t]
  \caption{Module $\aspmodule{col}$\label{asp:col}\label{asp:nonblocker}}
\begin{lstlisting}
collider(Y,X,Z) :- arrow(X,Y),arrow(Z,Y),X!=Y,var(X),var(Y),var(Z).
coll_desc(N,Y,X,Z) :- collider(Y,X,Z),dpath(Y,N).
nb(N,X,Y,S) :- in(N,S), collider(N,X,Y).
nb(N,X,Y,S) :- not in(N,S),not collider(N,X,Y),var(N),var(X),var(Y),set(S),N!=X,N!=Y,X!=Y.
nb(N,X,Y,S) :- not in(N,S),coll_desc(Z,N,X,Y),in(Z,S),var(N),var(X),var(Y).
\end{lstlisting}
\end{listing}
\begin{listing}[t]
  \caption{Module $\aspmodule{ap}(\Gpath,(v_i)_{i\leq k})$\label{asp:activepath}}
\begin{lstlisting}[escapeinside={(*}{*)}]
ap((*$v_1$*),(*$v_k$*),(*$\Gpath$*),S) :- (arrow((*$v_i$*),(*$v_{i+1}$*)))(*$_{i<k}$*), not in((*$v_1$*),S), not in((*$v_k$*),S), set(S), (nb((*$v_i$*),(*$v_{i-1}$*),(*$v_{i+1}$*),S))(*$_{1<i<k}$*).
 dep((*$v_1$*),(*$v_k$*),S) :- ap((*$v_1$*),(*$v_k$*),(*$\Gpath$*),S).
\end{lstlisting}
\end{listing}
This guarantees the `\emph{if}'-direction: if $x$ and $y$ are connected via a $\Zvar$-active path then they are dependent. 
For the `\emph{only if}'-direction, we require Module $\aspmodule{bp}(x,y,(\Gpath_i)_{i\leq k})$: for each pair of variables $x$ and $y$, we add the rule detailed in the listing to ensure that the absence of an active path between $x$ and $y$ implies independence between them; $(\Gpath_i)_{i\leq k}$ denotes the list of all paths between $x$ and $y$. The Module $\aspmodule{bp}$ in Listing~\ref{asp:activepath2} encodes the blocked path rules defined in Definition~\ref{def:causal ABAF: more rules}.
 The \textbf{indep}- and \textbf{dep}-predicates take two variables $x$, $y$, and a set $S$ of variables as arguments.
\begin{listing}[t]
  \caption{Module $\aspmodule{bp}(x,y,(\Gpath_i)_{i\leq k})$\label{asp:activepath2}}
\begin{lstlisting}[escapeinside={(*}{*)}]
indep(x,y,S) :- (not ap(x,y,(*$\Gpath_i$*),S))(*$_{i\leq k}$*), not in(x,S), not in(y,S), set(S).
\end{lstlisting}
\end{listing}
We note that, in general, the number of paths between two variables can be exponential (up to $\floor{(d-2)!e}$). To lower the number of paths, we make use of the observation that fixing independence facts amounts to removing edges between nodes.\footnote{This observation is key for constraint-based causal discovery algorithms such as PC~\citep{spirtes2000causation}.}
When fixing independence facts ($a\!\indep\!b\!\given\!\Xvar\!\gets$), we thus consider only the paths in the skeleton that do not contain $(a,b)$. 

As outlined in Proposition~\ref{prop:causal ABA add only facts}, fixing dependence facts requires only the addition of the blocked path rules corresponding to the fact. That is, adding the fact ($a\!\notindep\!b\!\given\!\Xvar\!$) only requires including the rules $\aspmodule{ap}(\Gpath,(v_i)_{i\leq k})$ and $\aspmodule{bp}(x,y,(\Gpath_i)_{i\leq k})$ to guarantee correctness. 

As discussed in \S\ref{sec:ABAF narrow def}, our proposed ABAF returns all the DAGs compatible with some fixed facts, representing relations amongst nodes, may these be conditional/marginal independencies and/or (un)directed causal relations (arrows and edges).
In the proposed instantiation of Causal ABA, ABA-PC, we input a set of facts in the form of independence relations and weight them according to their $p$-value.
\newcommand{\repl}{\textit{rep}}

\subsection{Sourcing Facts}\label{subsec:sourcing facts}

A DAG with $d$ nodes is fully characterised by $\frac{1}{2}d(d-1)2^{d-2}$ independence relations, growing exponentially in the number of nodes. Therefore, it is not computationally efficient to carry out all possible tests, as in \citep{hyttinen2014ASP}. 
Several solutions to this problem have been proposed in the Causal Discovery literature, e.g.,~\citep{spirtes2000causation,tsamardinos2006MMHC,colombo2014order}.
\cite{spirtes2000causation,tsamardinos2006MMHC,colombo2014order} all use conditional independence tests such as \citep{fisher1970statistical, zhang2011kci, Gretton2007hsic}. 
Other strategies to recover causal graphs from data, referred to as score-based methods, such as \citep{chickering2002ges, ramsey2017million} involve the use of statistical metrics that measure the added-value of adding/removing an arrow in terms of fit to the data. Hence they would return arrow weights. 
In this work, we use the MPC algorithm~\citep{colombo2014order}, which provably\footnote{under the assumptions of sufficiency (no unmeasured confounders), faithfulness (data represents a DAG) and perfect independence information, see \citep{colombo2014order} for detail and formal definitions. Note that the original PC strategy is based on the assumption that there will be no inconsistencies and therefore the algorithm does not test a pair of variables anymore once an independence is found. However, inconsistencies might arise when erroneous results are obtained.} recovers the underlying CPDAG from data, to source facts.
Let us illustrate the input facts we consider through our running example.

\begin{example}\label{ex:running_PC}
We run the MPC algorithm in Example~\ref{ex:intro} which performs 23 out of 24 tests, 
including the following.
\begin{align*}
    & r\!\indep\!wp\!\given\!\{ws\}  &&wp\!\indep\!ws\!\given\!\{r\}  &&r\!\indep\!wp\\
    & r\!\indep\!wp\!\given\!\{wr\}  &&wp\!\notindep\!ws\!\given\!\{r,wr\}\ &&r\!\indep\!wp\!\given\!\{wr,ws\}
\end{align*}
However, only $r\!\indep\!wp$ is correct; the only other independence $wp\!\indep_{\!\DAG}\,ws\!\given\!\{r,wr\}$ in $\DAG$ is wrongly classified. All other tests result in dependencies.

Based on this erroneous results, MPC yields the graph shown in Example~\ref{ex:intro} (right), deviating from the ground truth.
Crucially, the graph does not capture the independence relations listed above. 
In fact, there is no graph that satisfies the test results because it is not possible that $r$ and $wp$ are independent conditioned on any set, but $r$ and $ws$, as well as $wp$ and $ws$, are dependent. The dependencies indicate a path between $r$ and $wp$, leading to a contradiction.
\end{example}

Note that Algorithm~\ref{alg:ABAPC} is flexible to the choice of facts' source, e.g. we could have used the tests performed by \citep{tsamardinos2006MMHC} or, with a slight modification, the arrow weights from \citep{ramsey2017million}. 

\subsection{Weighting Facts}\label{subsec:fact selection}
Here we outline our strategy to weight independence tests results, based on their $p$-value and the size of the conditioning set. We use these weights as weak constraints and to rank facts and extensions.
As a result of using stable semantics, wrong tests can render empty extensions if they contradict another (set of) test(s).
Our aim is thus to exclude the wrong tests that create inconsistencies and cause our ABAF to output no extension. To this end, we define a simple heuristic to rank $p$-values from independence tests, given significance level $\alpha$, but insensitive to whether they fall below or above it. 
Firstly, we define the following normalising function:
\begin{equation*}
    \begin{split}
        &\gamma(p,\alpha) \!=\! 
        \begin{cases}  
            2p\alpha - 1              &\hspace{-0.2cm}\text{ iff } p < \alpha\\
            \frac{2\alpha-p-1}{2(\alpha-1)}  &\hspace{-0.2cm}\text{ otherwise } 
        \end{cases}
    \end{split}
\end{equation*}
Below is a plot of the function $\gamma$ across the $p$-value interval, for three commonly chosen levels of $\alpha$.
\begin{figure}[H]
    \centering
    \includegraphics[width=0.46\textwidth]{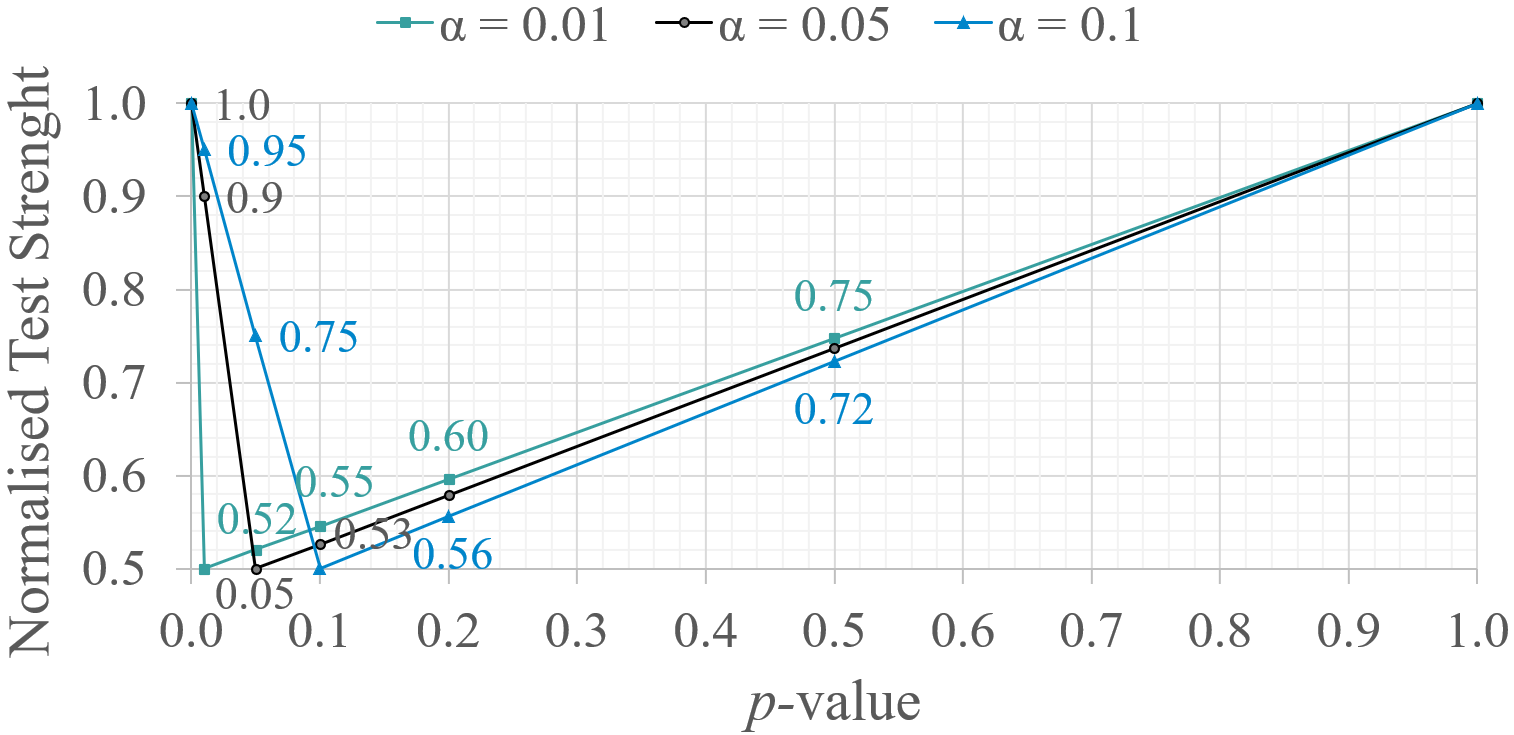}
    \label{fig:initialstrength}
\end{figure}

The output of $\gamma$, for a given $\alpha$ and $p$, follows the intuition that the most uncertainty is around the significance threshold $p=\alpha$ \citep{sellke2001calibration,berger2003could}, which we make correspond to the lowest value of $\gamma=0.5$. 

The final strength of the (in)dependence facts is obtained by weighting the output of the normalising function $\gamma$ by a factor penalising bigger sizes of the conditioning set \condSet:
\begin{equation}\label{eq:strength}
    \mathcal{S}(p,\alpha,s,d) = \frac{(1-s)}{(d-2)}\gamma(p,\alpha)
\end{equation}
where $s=|\condSet|$ the cardinality of the conditioning set and $d=|\nodeSet|$, the cardinality of the set of nodes in the graph.
The reason for weighting $\gamma$ by $s$ and $d$ follows the intuition that the accuracy of independence test lowers as the conditioning set size increases~\citep{sellke2001calibration}.

We use our final weights $\mathcal{S}$ to rank the test carried out by MPC. Then, our strategy is simple: exclude an incremental number of the lowest ranked tests until the returned extension is not empty. 
Let us illustrate our strategy. 
\begin{example}\label{ex:intro2}
 Consider again Example~\ref{ex:intro}. The results of the independence tests from MPC (using Fisher's Z~(\citeyear{fisher1970statistical}) and $\alpha=0.05$) have the following $p$-values (we show the same subset of the 23 tests carried out, as in Example~\ref{ex:running_PC}):
\begin{align*}
    &r\indep wp                    &&p=0.45  &&\mathcal{S}=0.71 \quad\\
    &r\indep wp\given \{ws\}        &&p=0.52  &&\mathcal{S}=0.37 \quad\\
    &r\indep wp \given \{wr\}       &&p=0.33  &&\mathcal{S}=0.32 \quad\\
    &wp\indep ws \given \{r\}       &&p=0.05  &&\mathcal{S}=0.25 \quad\\
    &r\indep wp \given \{wr,ws\}     &&p=0.39  &&\mathcal{S}=0.00 \quad\\
    &wp\notindep ws \given \{r,wr\}  &&p=0.03  &&\mathcal{S}=0.00
\end{align*}

We apply Eq.~\ref{eq:strength} to calculate $\mathcal{S}$. Ranking tests by $\mathcal{S}$, as shown above, the right test is the highest scoring one. 
Fixing all the tests returns no solution.
We thus start excluding the test with the lowest strength and progressively more until we find a model. In this example, the right DAG is obtained by excluding 9 of the performed tests, including the bottom five of the above list, and keeping the 14 strongest ones.
\end{example}

Here, we obtain exactly one DAG when excluding 40\% of the tests carried out by MPC. If we obtain multiple models, we score each of them as in Algorithm~\ref{alg:ABAPC}, lines 14-19.
In addition, we encode the (in)dependence facts as \emph{weak constraints}, treated as optimisation statements~\citep{GebserKS11}, to sort out sub-optimal extensions.

\begin{figure*}[h!t]
    \centering
    \includegraphics[width=0.98\textwidth]{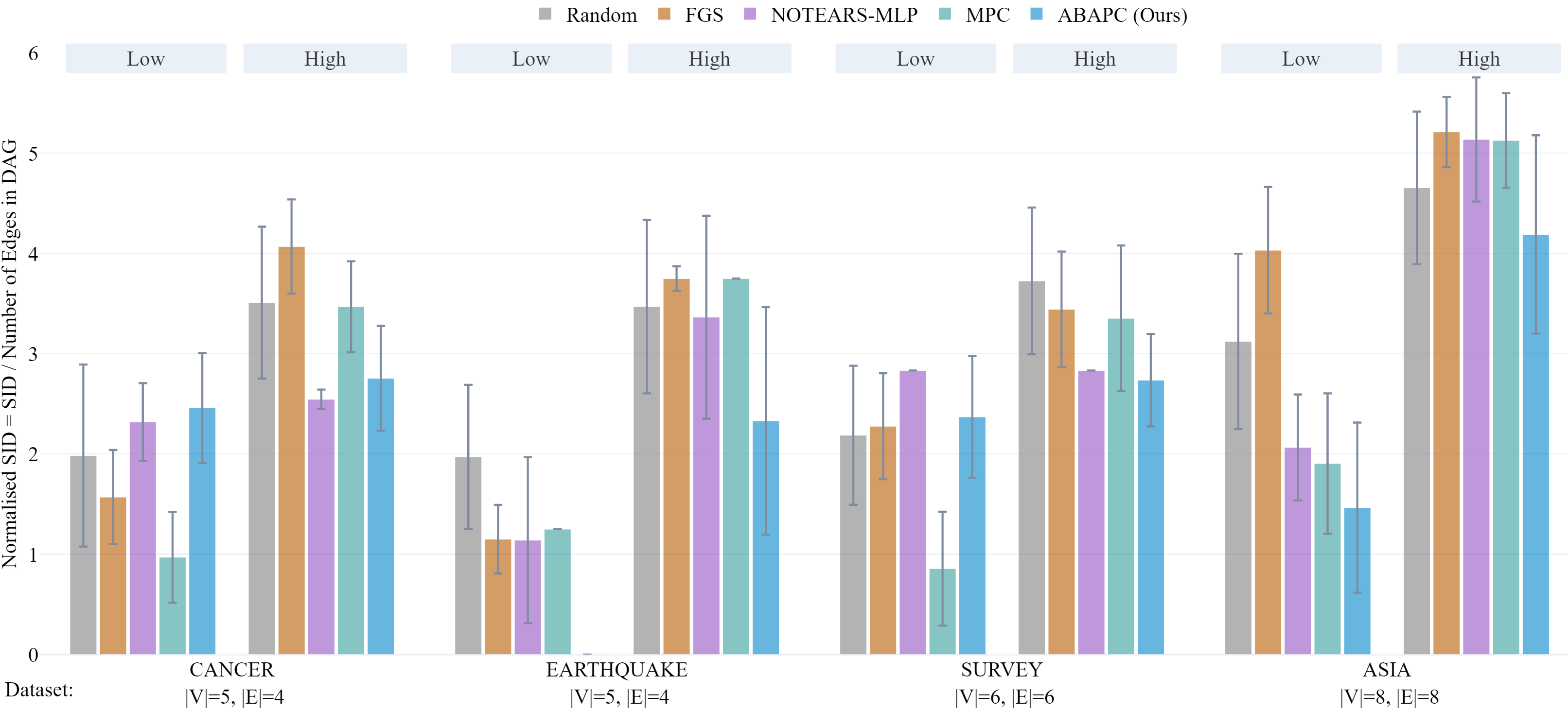}
    \caption{Normalised Structural Interventional Distance for four datasets from the \texttt{bnlearn} repository. Lower is better. Low (resp. High) is the SID for the best (resp. worst) DAG in the estimated CPDAG.}
    \label{fig:SID_cpdag}
\end{figure*}
Our weighting function is similar to the one proposed by \cite{DBLP:journals/jmlr/BrombergM09}, with two differences: we re-base around 0.5 instead of $1-\alpha$, to allow for more discrimination; and use the conditioning set size irrespective of the test's result,
instead of including it only in the case of dependence (trusting that $p$-values accurately reflect the probability of wrongly rejecting the null hypothesis).

We emphasize that classical independence tests are asymmetric in nature, and inference of dependence is only possible if there is enough evidence against the null hypothesis ($p\!<\!\alpha$) with an expected Type I error (rejecting independence when it is true) corresponding to $\alpha$. Conversely, no inference is possible when the $p\!\geq\!\alpha$, when the null hypothesis cannot be rejected, since $p$-values are distributed uniformily in $[0,1]$ under \nullH. 

As pointed out in \citep{DBLP:journals/jmlr/BrombergM09,hyttinen2014ASP}, using $p$-values directly as strength is common but neither sound nor consistent. Transforming $p$-values to probability estimates, 
e.g. as in \citep{ jabbari2017discovery,Claassen2012bayesian,triantafillou2014learning}, would address this point, but out of scope for this work.
A possible alternative to weighting and excluding facts 
might also be the use of less strict ABA semantics, left out of our experiments since not available in the ASP implementation used.

\section{Empirical Evaluation}\label{sec:experiments}
We evaluate our ABA-PC algorithm on four datasets from the \texttt{bnlearn} repository~\citep{Scutari14bnlearn}, which hosts commonly used benchmarks in Causal Discovery, some of which based on real published experiments or expert opinions. 
We use the Asia, Cancer, Earthquake and Survey datasets, which represent problems of decision making in the medical, law and policy domains (see Appendix \S\ref{sec:bnlearn} for details).
Implementation and computing infrastructure details, including code to reproduce the experiments, are in Appendix \S\ref{sec:implem_det}.

\subsubsection{Evaluation Metrics and Baselines}
For evaluation, we use a prominent metric in causal discovery: Structural Interventional Distance (SID)~\citep{peters2015structural} measures the deviation in the causal effects estimation deriving from a mistake in the estimated graph. 
SID works as a ``downstream task" error rate for the causal inference task, which has causal graphs as a pre-requisite. 
We calculate SID between the estimated and the true CPDAG
and repeat the experiments 50 times per dataset to record confidence intervals. Given that a CPDAG is a mixed graph, SID is calculated for the worst and best scenarios. 
In order to compare across graphs 
with different number of edges, 
we normalise SID (NSID) dividing it by the number of edges in the true DAG.
NSID can go above 100\% since extra edges could be introduced in the structure. We provide details on the metrics in Appendix \S\ref{sec:metrics_det} and  results based on additional metrics (SHD, F1 score, precision and recall) in Appendix \S\ref{sec:add_results}.

We compare ABA-PC to four baselines: a Random sample of graphs of the right dimensions (\nodeSet,\edgSet); 
Fast Greedy Search (FGS)~\citep{ramsey2017million} and NOTEARS-MLP~\citep{zheng2020notears_mlp} which use, respectively, the Bayesian Information Criterion and Multilayer Perceptrons with a continuous formulation of acyclicity to optimise the  graph's fit to the data; and MPC~\citep{colombo2014order}.\footnote{We would have liked to compare to the method closest to our work, i.e. \citep{DBLP:journals/jmlr/BrombergM09} but unfortunately there is no implementation available.}
More details are provided in Appendix \S\ref{sec:baseline_det}.

\subsubsection{Results} The results of our experiments are in Fig.~\ref{fig:SID_cpdag}. 
Best and worst SID are in the (Low, resp. High) sections for each dataset;
the number of edges and nodes in each dataset is 
given below the x-axis labels. 
ABA-PC ranks $1^{st}$ in the worst case SID (High) for all datasets. 
It performs significantly (w.r.t.\ t-tests of difference in means, see \S\ref{sec:stats}) better than all baselines on three out of four datasets (Cancer, Earhquake and Asia) and is on par with NOTEARS-MLP for the Survey data. Furthermore, ABA-PC performs significantly better than MPC for all datasets.
This demonstrates how, with the same underlying information from the data, 
our proposed method returns more accurate CPDAGs in the worst case scenario. For the best case SID (Low), ABA-PC is significantly better than all baselines for two datasets (Earthquake and Asia). 
Overall, we observe that ABA-PC performs well on benchmark data compared to a varied selection of baselines from the literature.
\begin{figure}[t]
    \centering
    \includegraphics[width=0.46\textwidth]{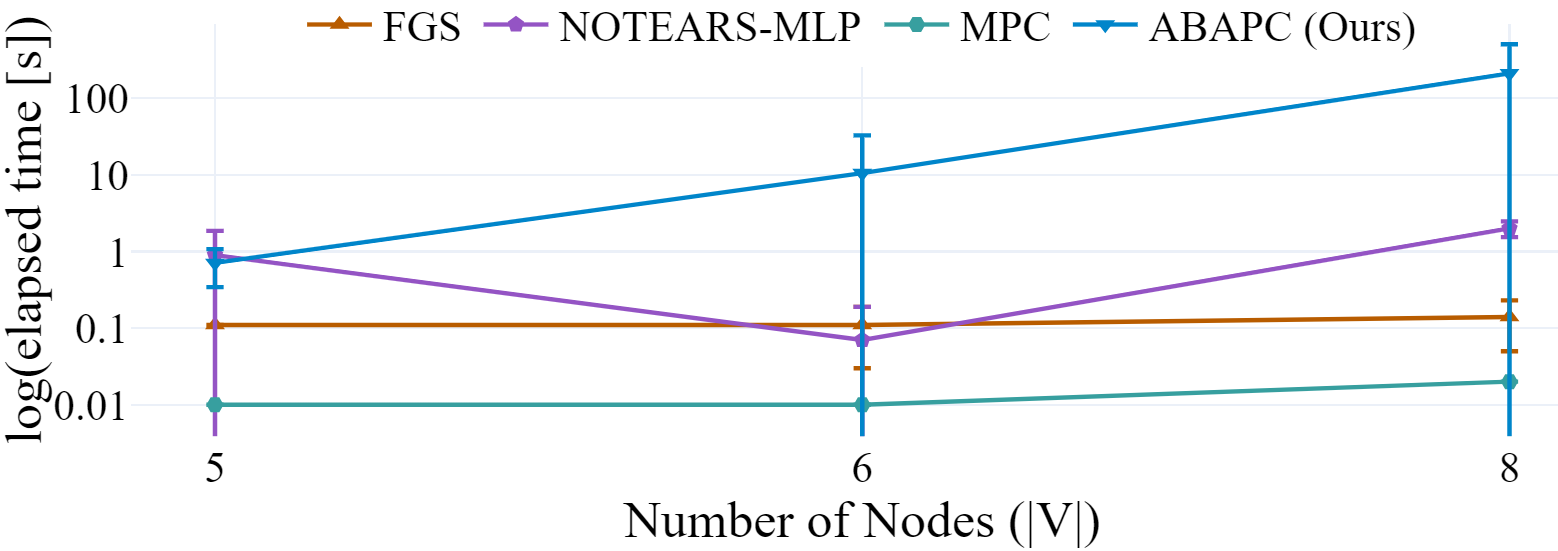}
    \caption{Mean and Standard Deviation of the elapsed time in log scale by number of nodes averaged over 50 runs.}
    \label{fig:runtime}
\end{figure}
\subsubsection{Scalability} In Fig.~\ref{fig:runtime} we show the elapsed time (on a log scale) by the number of nodes. This are the recorded times for the experiments in Fig.~\ref{fig:SID_cpdag} with 50 repetitions per dataset. As we can see, ABA-PC is the least efficient method. 
The main reasons for this are the complexity of both the grounding of logical rules and the calculation of the extensions, which clingo carries out exactly and efficiently, but still constitute a bottleneck.
We already identified 
avenues of future work, discussed next, to address scaling limitations of our implementation, given the promising results shown in Fig.~\ref{fig:SID_cpdag}.

 \section{Conclusion}

 We proposed a novel argumentation-based approach to Causal Discovery, targeting the resolution of inconsistencies in data, and showed that it outperforms existing statistics-based methods on four (standard) datasets. Our approach uses independence tests and their $p$-values to narrow down DAGs most fitting to the data, drawn from stable extensions of ABA frameworks.
Other methods to identify and resolve inconsistencies in data for causal discovery have been proposed, e.g. by~\cite{ramsey2006CPC,colombo2014order}, but they focus on marking orientations as ambiguous in the presence of inconsistencies, rather than actually resolving the inconsistencies as we do.

Our proposed framework allows for the introduction of weighted arrows and edges, on top of independencies, which would allow to integrate, as future work, other data-centric methods like score-based causal discovery algorithms (e.g. \citep{chickering2002ges, ramsey2017million, Claassen2012bayesian}.
As for the scalability, we cannot process more than 10 variables at the current state. We are currently working on making the processing of extensions more efficient and on incremental solving
to avoid re-grounding when deleting independence facts.
Additionally, we would like to extend our approach to deal with latent confounders, in line with \citep{Colombo2012RFCI, hyttinen2014ASP} and cycles \citep{DBLP:journals/ijar/RantanenHJ20,Richardson1996CCD,hyttinen2014ASP} and experiment with other argumentation semantics in the literature, making use of a recently developed solver for non-flat ABA~\citep{DBLP:journals/corr/abs-2404-11431}.
Finally, we plan to explore  the explainability capabilities intrinsic in an ABA framework~\citep{CyrasFST2018}, which we believe may bring 
great value to causal discovery
in a collaborative human-AI discovery process~\citep{contestable_ecai}.





\section*{Acknowledgemets}
Russo was supported by UK Research and Innovation (grant number EP/S023356/1), in the UKRI Centre for Doctoral Training in Safe and Trusted Artificial Intelligence (\url{www.safeandtrustedai.org}). Rapberger and Toni were funded by the ERC under the EU’s Horizon 2020 research and innovation programme (grant number 101020934) and Toni also by J.P. Morgan and by the Royal Academy of Engineering under the Research Chairs and Senior Research Fellowships scheme. 

\bibliographystyle{kr}
\bibliography{bib}

\appendix
\clearpage
\section{Proofs of Section~\ref{sec:causal ABA}}\label{sec:proof}

\PropDAGABA*

\begin{proof} 
By the well-known relations between the semantics~\citep{BaroniCG18}, i.e., $\cf(D)\supseteq \adm(D)\supseteq \com(D)\supseteq \prf(D) \supseteq \stb(D)$ and $\cf(D)\supseteq \stb(D)$, it suffices to prove the statement for conflict-free and stable semantics. 
\begin{itemize}
    \item ($\sigma=\cf$) $(\subseteq)$ Let $S\in \cf(D_\dagg)$. By definition, $S$ cannot contain cycles; hence, $S$ corresponds to a DAG. 

$(\supseteq)$ Let $G=(V,E)$ be a DAG. Let $S=\{\arrow_{x,y}\mid (x,y)\in E\}$. By acyclicity of $G$, we obtain that $S$ is conflict-free.
    \item ($\sigma=\stb$) $(\subseteq)$ Let $S\in \stb(D_\dagg)$. By definition, $S$ cannot contain cycles; hence, $S$ corresponds to a DAG. 

$(\supseteq)$ Let $G=(V,E)$ be a DAG. Let $S=\{\arrow_{x,y}\mid (x,y)\in E\}\cup \{\noedge_{xy}\mid (x,y),(y,x)\notin E\}$. Note that $S$ contains exactly one of $\arrow_{xy}$, $\arrow_{yx}$, $\noedge_{xy}$ (since $G$ does not contain a bi-directed arrow and $S$ contains $\noedge_{xy}$ only if $x$ and $y$ are not linked in $G$). Therefore, $S$ attacks each assumption which is not contained in $S$. Moreover, $S$ does not contain cycles by acyclicity of $G$. Therefore, $S$ is conflict-free. \qedhere
\end{itemize}
\end{proof}

\LemCausalABAStablecoincide*
\begin{proof}
    Since $\prf(D) \supseteq \stb(D)$ it suffices to show $\prf(D_\dagg)\subseteq \stb(D_\dagg)$. 
    Let $S\in\prf(D_\dagg)$.
    The main observation is that for each pair of variables $x,y\in V$, either $\noedge_{xy}$, $\arrow_{xy}$ or $\arrow_{yx}$ is contained in $S$. Towards a contradiction, suppose there exists a pair of variables such that $S\cap \{\noedge_{xy},\arrow_{xy},\arrow_{yx}\}=\emptyset$. It is easy to see that $S\cup \{\noedge_{xy}\}$ is admissible ($\noedge_{xy}$ is only attacked by the corresponding arrows and defends itself against these attacks); contradiction to $S$ being a $\subseteq$-maximal admissible set. We obtain $\prf(D_\dagg) = \stb(D_\dagg)$.
\end{proof}

Below, we make use of the following definition.\begin{definition}
For a set of assumptions $S$, we let $\DAG(S)=S\cap \mathcal{A}_\arrow$ denote the graph corresponding to $S$.     
\end{definition}

\PropCausalABACorrectness*
\begin{proof}
By Lemma~\ref{lem:ABAcausal stable pref etc coincide} it suffices to prove the proposition for stable semantics. 

    $(\Rightarrow)$ Suppose  $x\indep y\given \Zvar\in S$. 
    Towards a contradiction, suppose $x\notindep y\given \Zvar$ in $G(S)$.  
 Then there exists a $\Zvar$-active path $\Gpath$ between $x$ and $y$ in $G(S)$. Then there is a $\Zvar$-active x-y-collider-tree $\tree$ where $\Gpath$ is the connecting path between $x$ and $y$ (i.e., $\Gpath=\ctpath$). By definition of the graph $G(S)$, each directed arrow in $\Gpath$ is contained in $S$. Hence, $S\vdash \contrary{x\indep y\given \Zvar}$ can be derived, contradiction to conflict-freeness of $S$. 

    $(\Leftarrow)$ For the other direction, suppose $x\indep y\given \Zvar$ in $G(S)$.
    We show that each set of assumptions $T$ with $T\vdash \contrary{x\indep y\given \Zvar}$ is attacked.
    Consider a $\Zvar$-active x-y-collider-tree $\tree$. Let $\ctpath$ denote the path corresponding to $\tree$.
    Note that $\ctpath\notin S$, otherwise, $x$ and $y$ are d-connected given $\Zvar$. 
    Let $(a,b)\in \ctpath$ denote the arrow which is not contained in $S$. Since $S$ is stable, it holds that $\{\arrow_{ab},\arrow_{ba},\noedge_{ab}\}\cap S=\emptyset$. Therefore, either $\noedge_{ab}\in S$ or $\arrow_{ba}\in S$. 
    Therefore, $\ctpath$ is attacked by $S$. Since $\tree$ was arbitrary, we obtain $x\indep y\given \Zvar$ is defended by $S$ and therefore $x\indep y\given \Zvar\in S$.    
\end{proof}

\PropCausalABAforAdmComCf*
\begin{proof}
 Suppose  $x\indep y\given \Zvar\in S$. 
    Towards a contradiction, suppose $x\notindep y\given \Zvar$ in $G(S)$. Then there exists a $\Zvar$-active path $\Gpath$ between $x$ and $y$ in $G(S)$. Then there is a $\Zvar$-active x-y-collider-tree $\tree$ where $\Gpath=\ctpath$ is the connecting path between $x$ and $y$. By definition of the graph $G(S)$, each directed arrow in $\ctpath$ is contained in $S$. Hence, $S\vdash \contrary{x\indep y\given \Zvar}$ can be derived, contradiction to conflict-freeness of $S$.  
\end{proof} 

\PropIntegrateDirectedEdges*
\begin{proof}
The proof is analogous to the proof of Proposition~\ref{prop:ABA Dsep semantics correspondence}. The rule $r$ ensures that $S$ contains the assumption $x\indep y\given \Zvar$ resp.\ $\arrow_{xy}$.   
\end{proof}

\PropCABAsinglestatement*
\begin{proof}
By Lemma~\ref{lem:ABAcausal stable pref etc coincide} it suffices to prove the proposition for stable semantics. 
Let $r=\contrary{x\indep y\given \Zvar}\gets$.

    $(\Rightarrow)$ This direction is analogous to the proof of Proposition~\ref{prop:ABA Dsep semantics correspondence}.

    $(\Leftarrow)$ For the other direction, suppose $a\indep b\given \Xvar$ in $G(S)$.
    We proceed by case distinction. 
    
    \underline{Case 1 ($a=x$, $b=y$, $\Xvar=\Zvar$)}

    Towards a contradiction, suppose $x\indep y\given\Zvar\notin S$. Since $S$ is closed, there is some x-y-path $\Gpath$ such that $\blockedpath_{\Gpath\mid \Zvar}\notin S$ (otherwise, $S\vdash x\indep y\given\Zvar$, contradiction).
    Therefore, $S$ attacks $\blockedpath_{\Gpath\mid \Zvar}$. 
    This is the case if $S$ derives $\activepath_{\Gpath\mid \Zvar}$. By definition of $D_\dsepbp$, this is the case if $S$ contains an $\Zvar$-active x-y-collider-tree $\tree$.
    Consequently, $x$ and $y$ are d-connected given $\Zvar$, contradiction to $x\indep y\given \Zvar$ in $G(S)$.

    \underline{Case 2 ($a\neq x$ or $b\neq y$ or  $\Xvar\neq \Zvar$)} 

    Analogous to the proof of Proposition~\ref{prop:ABA Dsep semantics correspondence}.
\end{proof}

\section{ASP Encodings: Additional Details}\label{sec:asp_det}
We recall the standard translation from ABA to LP. We assume that the ABA has precisely one contrary for each assumption. Given an ABAF $D=(\mathcal{L},\mathcal{A},\mathcal{R},\contraryempty)$ and an atom $p\in \mathcal L$, we let 
 $$\repl(p)=
\begin{cases}
\textit{not}\ \contrary{p}, & \text{ if }p\in \mathcal{A}\\
\contrary{a}, & \text{ if }p = \contrary{a}\in\contrary{\mathcal{A}}.
\end{cases}$$
We extend the operator to ABA rules element-wise: 
$\repl(r)=\repl(head(r))\gets \{\repl(p)\mid r\in body(r)\}$.
    For an LP-ABAF $D\!=\!(\mathcal{L},\mathcal{R},\mathcal{A},\contraryempty)$,
    we define the \emph{associated LP} $P_D\!=\!\{\repl(r)\!\mid\! r\!\in\! \mathcal{R}\}$.

\begin{listing}[t]
  \caption{Module $\aspmodule{dag}$\label{asp:dag}}
\begin{lstlisting}
arrow(X,Y)|arrow(Y,X)|not edge(X,Y) :- var(X),var(Y), X!=Y.
edge(X,Y):- arrow(X,Y), X!=Y, var(X), var(Y).
edge(X,Y):- edge(Y,X), X!=Y, var(X), var(Y).
:- arrow(X,Y), not edge(X,Y), var(X), var(Y).
:- edge(X,Y), not arrow(X,Y), not arrow(Y,X), var(X), var(Y).
:- arrow(Y,X), arrow(X,Y), X!=Y, var(X), var(Y).
dpath(X,Y):- arrow(X,Y), X!=Y, var(X), var(Y).
dpath(X,Y):- arrow(X,Z), dpath(Z,Y), var(X). 
:- dpath(X,X), var(X).
\end{lstlisting}
\end{listing}
Following the standard ABA to LP translation, each ABA atom $\arrow_{xy}$ is translated to $\textit{not}\ \contrary{\arrow_{xy}}$. 
With a slight deviation from the causal ABAF, we identify ``not $\arrow_{yx}$'' simply with \textbf{arrow}(x,y) and ``not $\noedge_{xy}$'' with \textbf{edge}(x,y). Listing~\ref{asp:dag} contains our DAG encoding (as expected, \textbf{var} specifies variables). 
In Line~1, we guess one an arrow or the absence of an arrow; the remaining lines enforce that the graph is acyclic. 
The code faithfully captures the basis of our causal ABAF: each answer set of Module $\Pi_{dag}$ in Listing~\ref{asp:dag} corresponds to precisely one DAG of size $|V|$ for a given set $V$ of variables.

\section{Experimental Details}\label{sec:exp_det}

\subsection{\texttt{bnlearn} datasets}\label{sec:bnlearn}
\setlength\intextsep{0pt}
For our empirical evaluation (see \S\ref{sec:experiments}), we used four datasets from the \texttt{bnlearn} repository.\footnote{\url{https://www.bnlearn.com/bnrepository/}} This repository is widely used for research in Causal Discovery and hosts a number of commonly used benchmarks, some of which result from real published experiments or from the collection of expert opinions on the causal graph and the conditional probability tables necessary to create a Bayesian Network. 
Specifically, we use the Asia, Cancer, Earthquake and Survey datasets, as reported in reported in Table~\ref{tab:real_data_det}, which represent problems of decision making in the medical, law and policy domains. The datasets are from the small categories with number of nodes varying from 5 to 8. Details on the number of nodes, edges and density of the DAGs underlying the Bayesian Networks, together with links to a more detailed description on the \texttt{bnlearn} repository can be found in Table~\ref{tab:real_data_det}.
\vspace{0.3cm}
\begin{table}[h!t]
    \caption{Details of dataset from \texttt{bnlearn}.}
    \label{tab:real_data_det}    
    \centering
    \begin{tabular}{r|ccc}
Dataset Name & $|N|$ &  $|E|$  & $|E|/|N|$ \\
\hline
\href{https://www.bnlearn.com/bnrepository/discrete-small.html#cancer}{CANCER} & 5 & 4 & 0.8\\
\href{https://www.bnlearn.com/bnrepository/discrete-small.html#earthquake}{EARTHQUAKE} & 5 & 4  & 0.8\\
\href{https://www.bnlearn.com/bnrepository/discrete-small.html#survey}{SURVEY} & 6 & 6 & 1\\
\href{https://www.bnlearn.com/bnrepository/discrete-small.html#asia}{ASIA} & 8 & 8  & 1\\
    \end{tabular}
\end{table}

Having downloaded all the .bif files from the repository, we load the Bayesian network and the associated conditional probability tables and sample 5000 observations with 50 different seeds to measure variance and confidence intervals.\footnote{We also run all the experiments shown with 2000 samples and resulted in analogous results, hence omitted.}

\subsection{Implementation Details}\label{sec:implem_det}
We provide an implementation of ABA-PC using clingo~\citep{GebserKKS17} version 5.6.2 and python 3.10. The code is available at the following repository: 
\url{https://github.com/briziorusso/ArgCausalDisco}.
In the repository, we also made available the code to reproduce all experiments and we saved all the plots, presented herein and in the main text, in HTML format.\footnote{\url{https://github.com/briziorusso/ArgCausalDisco/tree/public/results/figs}} Downloading and opening them in a browser allows the inspection of all the numbers behind the plots in an interactive way. 

\subsubsection{Hyperparameters} We used default parameters for all the methods. For MPC and ABAPC (ours) we used Fisher Z test \citep{fisher1970statistical}, as implemented in \texttt{causal-learn}, with significance threshold $\alpha=0.05$.

\subsubsection{Computing infrastructure} 
Our proposed method, together with MPC and FGS do not benefit from GPU accelleration. All the results were ran on Intel(R) Xeon(R) w5-2455X CPU with 4600 max MHz and 128GB of RAM. 
The NOTEARS-MLP was ran on NVIDIA(R) GeForce RTX 4090 GPU with 24GB dedicated RAM.

\subsection{Evaluation Metrics}\label{sec:metrics_det}
We evaluated the estimated graphs with five commonly used metrics in causal discovery (as in e.g. \citep{zheng2020notears_mlp,Lachapelle2020grandag,ramsey2017million,colombo2014order}):
\begin{itemize}
    \item Structural Intervention Distance (SID) 
    \item Structural Hamming Distance (SHD) = E + M + R
    \item Precision = TP/(TP + FP)
    \item Recall = TP/(TP + FN)
    \item F1 = $2\times$Precision*Recall/(Precision+Recall)
\end{itemize}

SID was proposed in \citep{peters2015structural} and quantifies the agreement to a causal graph in terms of interventional distributions. It aims at quantifying the incorrect causal inference estimations stemming out of a mistake in the causal graph estimation, akin to a downstream task error on a pre-processing step.
SHD is a graphical metric that counts the number of mistakes present in an estimated directed graph compared to a ground truth one. In particular it sums the extra edges, the missing ones and the wrong orientations. The lower the SHD the better. In the formula, Extra (E) is the set of extra edges and Missing (M) are the ones missing from the skeleton of the estimated graph. Reversed (R) are directed edges with incorrect direction.
Precision and Recall measure the proportion of correct orientations based on the estimated graph and the true one, respectively. In the formulae, True Positive (TP) is the number of estimated edges with correct direction; False Positive (FP) is an edge which is not in the skeleton of the true graph. True Negative (TN) and False Negative (FN) are edges that are not in the true graph and correctly (resp, incorrectly) removed from the estimated graph. 
Finally, F1 score is the harmonic mean of precision and recall.

We carried out the evaluation in the main text on CPDAGs. As discussed in \S\ref{sec:prelim}, CPDAGs represent Markov Equivalence Classes. MECs are all that can be inferred from a given set of independence relations, since multiple DAGs can entail the same set of independencies. Both our method and NOTEARS-MLP output DAGs rather than CPDAGs (which are the output of the other two baselines used, FGS and MPC). For the methods returning DAGs, we first transform the DAG to a CPDAG (isolating skeleton and v-structures) and then calculate the metrics presented above. Evaluation on DAGs is provided in \S\ref{sec:add_results} for completeness.
\begin{figure*}[ht]
    \centering
    \includegraphics[width=\textwidth]{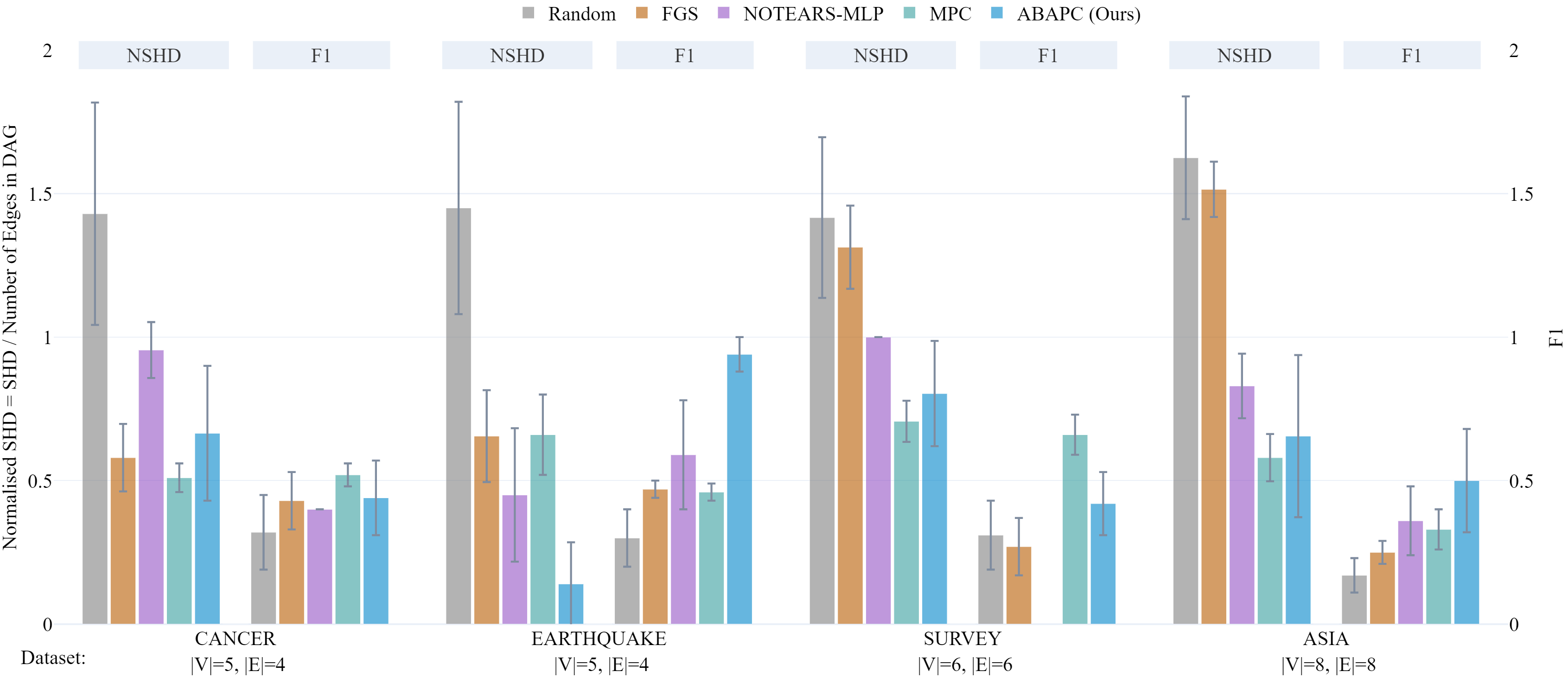}
    \caption{Normalised Structural Hamming Distance (SHD, left y-axis) and F1 score (right y-axis) for the estimated CPDAGs for four datasets in the \texttt{bnlearn} repository. Lower is better for NSHD and higher is better for F1.}
    \label{fig:SHD_F1}
\end{figure*}
\subsection{Baselines}\label{sec:baseline_det}
We used the following four baselines with respective implementations (see Section~\ref{sec:experiments} for context):
\begin{itemize}
    \item A Random baseline (RND) as in \citep{Lachapelle2020grandag}, by just sampling 10 random graphs with the same number of nodes and edges as the ground truth. 
    \item Majority-PC\footnote{ \url{https://github.com/briziorusso/ArgCausalDisco/blob/public/cd_algorithms/PC.py}} (MPC)~\citep{colombo2014order} is a constraint-based Causal Discovery algorithm. It uses independence tests and graphical rules based on d-separation to extract a CPDAG from the data. MPC is an improved version of the original Peter-Clark (PC) algorithm~\citep{spirtes2000causation} that renders it order-independent while mantaining soundness and completeness with infinite data.
    \item Fast Greedy equivalence Search\footnote{\url{https://github.com/bd2kccd/py-causal}} (FGS)~\citep{ramsey2017million} is a score-based Causal Discovery algorithm. It is a fast implementation of GES \citep{chickering2002ges} where graphs are evaluated using the Bayesian Information Criterion (BIC) upon addition or deletion of an edge, in a greedy fashion, involving the evaluation of insertion and removal of edges in a forward and backward fashion. Its output is a CPDAG.
    \item NOTEARS-MLP\footnote{\url{https://github.com/xunzheng/notears}} (NT)~\citep{zheng2020notears_mlp} learns a non-linear SEM via continuous optimisation. Having a Multi-Layer Perceptron (MLP) at its core, this method should adapt to different functional dependencies among the variables. The optimisation is carried out via augmented Lagrangian with a continuous formulation of acyclicity \citep{zheng2018notears}, outputting a DAG.
\end{itemize}

\subsection{Additional Results}\label{sec:add_results}

Here we provide results that complement the ones in \S\ref{sec:experiments} in the main text. Specifically, we evaluate our method based on SHD, F1 score, Precision and Recall (see \S\ref{sec:metrics_det} for details).

\subsubsection{Additional Metrics}

\begin{figure*}[t]
    \centering
    \includegraphics[width=\textwidth]{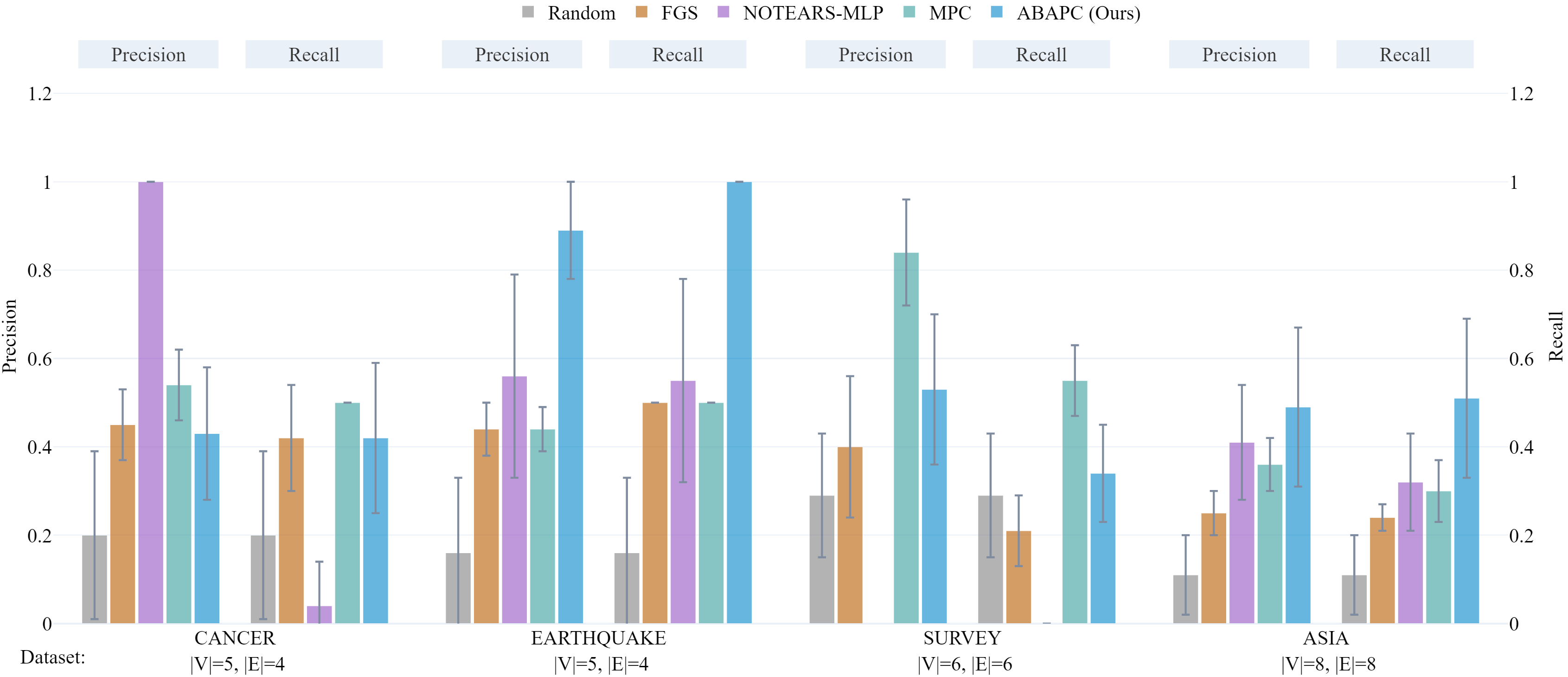}
    \caption{Precision (left y-axis) and Recall (right y-axis) for the estimated CPDAGs for four datasets in the \texttt{bnlearn} repository. Higher is better for both metrics.}
    \label{fig:pre_rec}
\end{figure*}

\begin{figure*}[t]
    \centering
    \includegraphics[width=\textwidth]{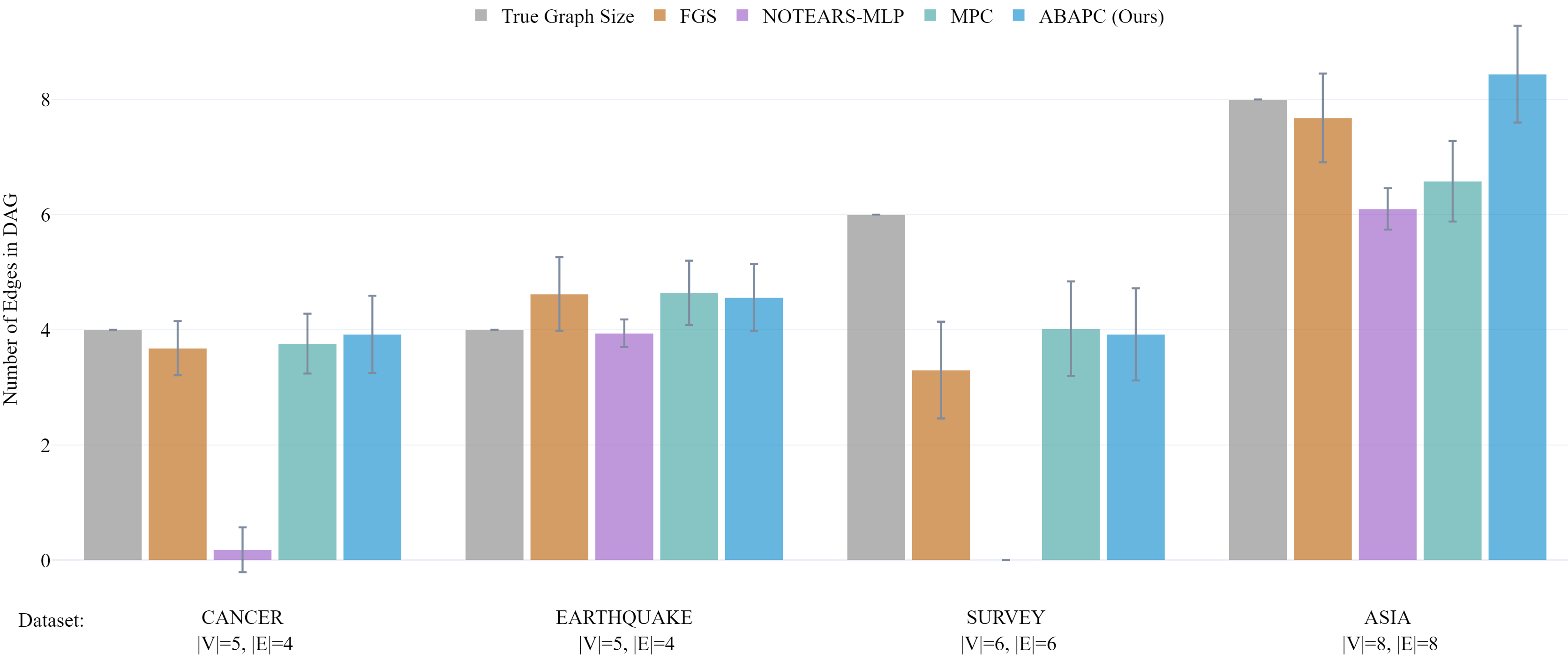}
    \caption{Number of edges in the estimated CPDAGs compared to ground truth.}
    \label{fig:ESG}
\end{figure*}

Additionally to the results shown in Fig.~\ref{fig:SID_cpdag} in the main text, we evaluate our proposed method with four other metrics commonly used in Causal Discovery: SHD, F1 (Fig.~\ref{fig:SHD_F1}), Precision and Recall (Fig.~\ref{fig:pre_rec}). Furthermore, we show the average size of the estimated CPDAGs in (Fig.~\ref{fig:ESG}). From Fig.~\ref{fig:SHD_F1}, we can see that, according to Normalised SHD, ABA-PC is the best method for the Earthquake dataset and not significantly different from MPC (ranking first) for the other datasets. In terms of F1 score, ABA-PC is significantly better than all baselines for the Earthquake and Asia datasets, on par with all the others for Cancer and ranking second, after MPC, for the Survey dataset. Precision and Recall provide a breakdown of the F1 score (which is their harmonic mean), hence follows similar patterns. From the estimated graph size (the number of edges in the estimated graph) in Fig.~\ref{fig:ESG} we can observe that ABA-PC is generally in line with the true graph size, apart from the Survey dataset for which some of the edges are missed.

\subsubsection{DAGs Evaluation}
\begin{figure*}
    \centering
    \includegraphics[width=\textwidth]{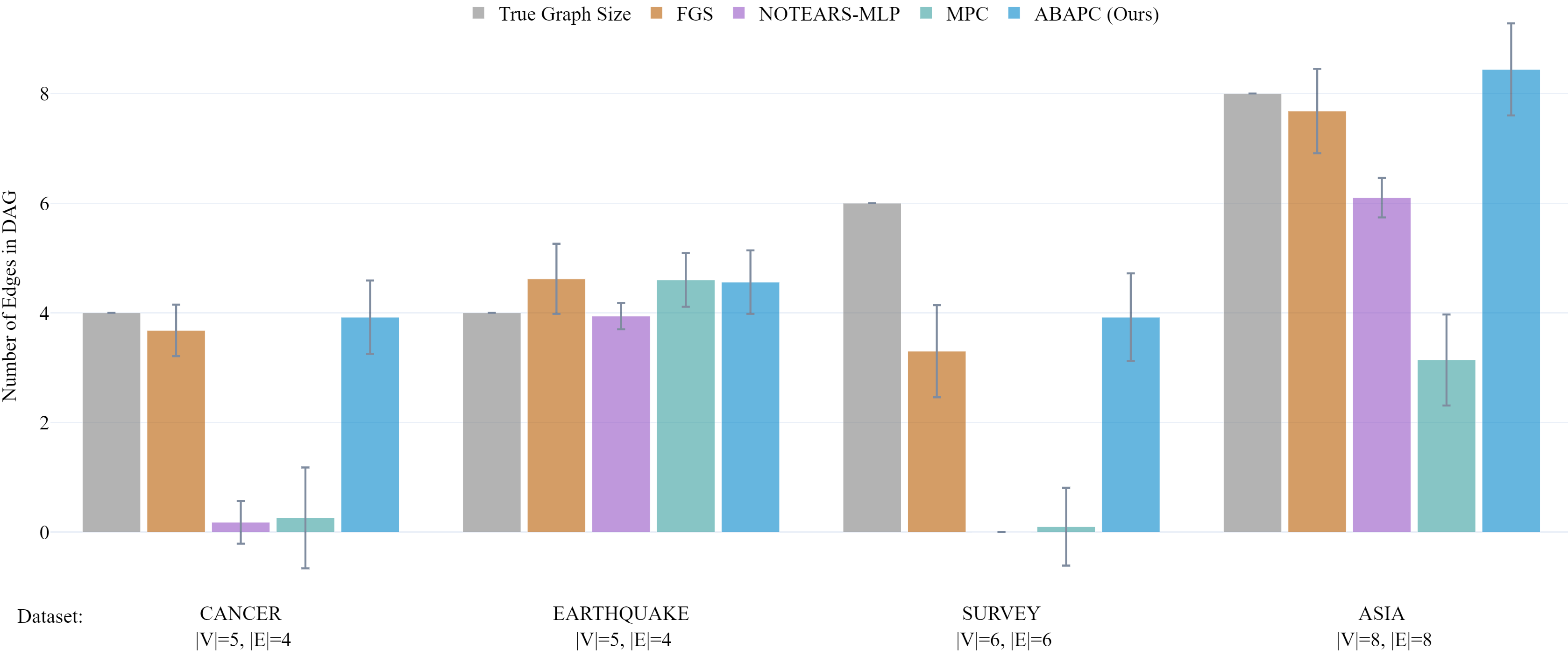}
    \caption{Number of edges in the estimated DAGs compared to ground truth.}
    \label{fig:ESG_DAG}
\end{figure*}

In the main text, we evaluated our method based on the estimated CPDAG compared to the ground truth one. Here, for completeness, we report results based on DAGs. Indeed, both our method and NOTEARS-MLP have DAGs in output. Note that this evaluation penalises the methods outputting CPDAGs. In transforming CPDAGs to DAGs, we had to only select the directed arrows in order to extract DAGs from the output CPDAGs (see Fig.~\ref{fig:ESG} and~\ref{fig:ESG_DAG} for the average size of the estimated CPDAGs and DAGs, respectively).

\begin{figure*}
    \centering
    \includegraphics[width=\textwidth]{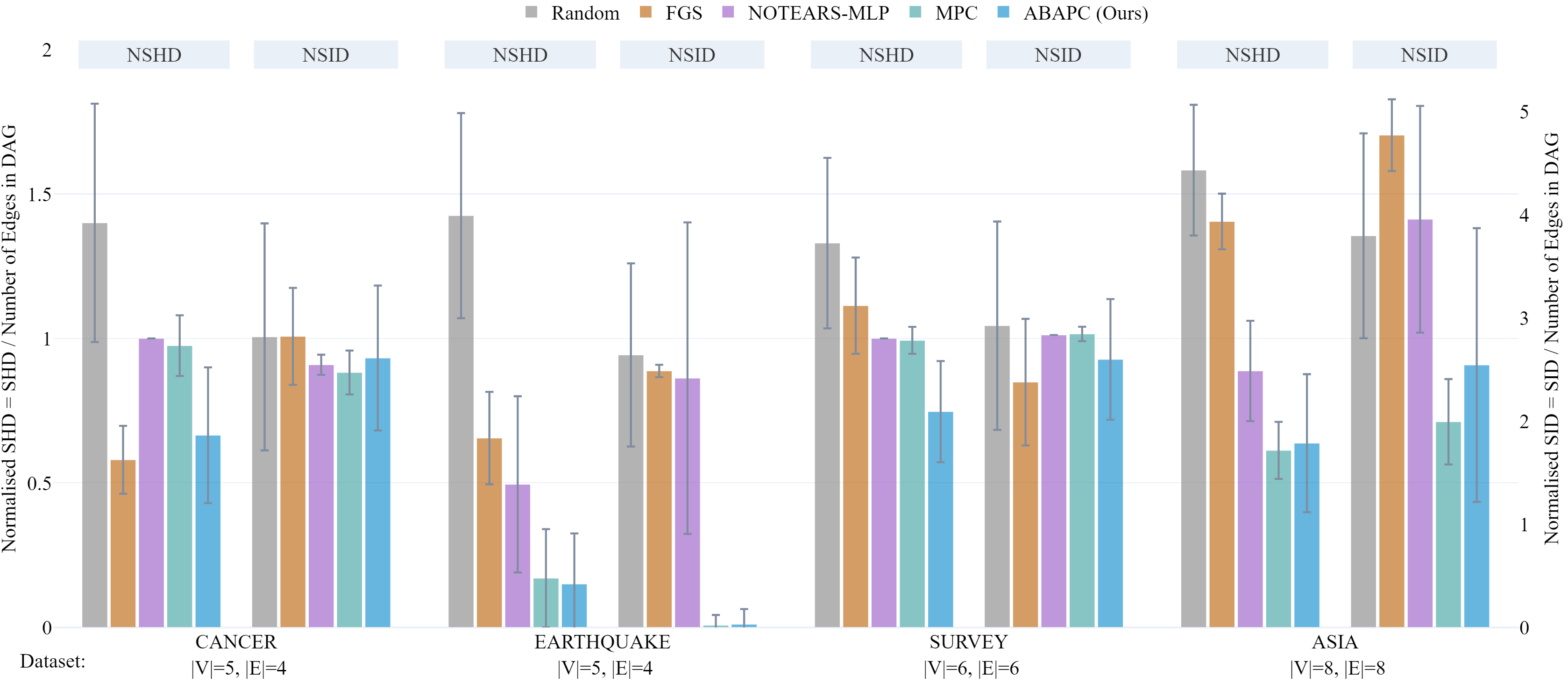}
    \caption{Normalised Structural Hamming Distance (SHD, left y-axis) and Structural Intervention Distance (SID, right y-axis) for the estimated DAGs for four datasets in the \texttt{bnlearn} repository. Lower is better for both metrics.}
    \label{fig:SHD_SID_DAG}
\end{figure*}

\begin{figure*}
    \centering
    \includegraphics[width=\textwidth]{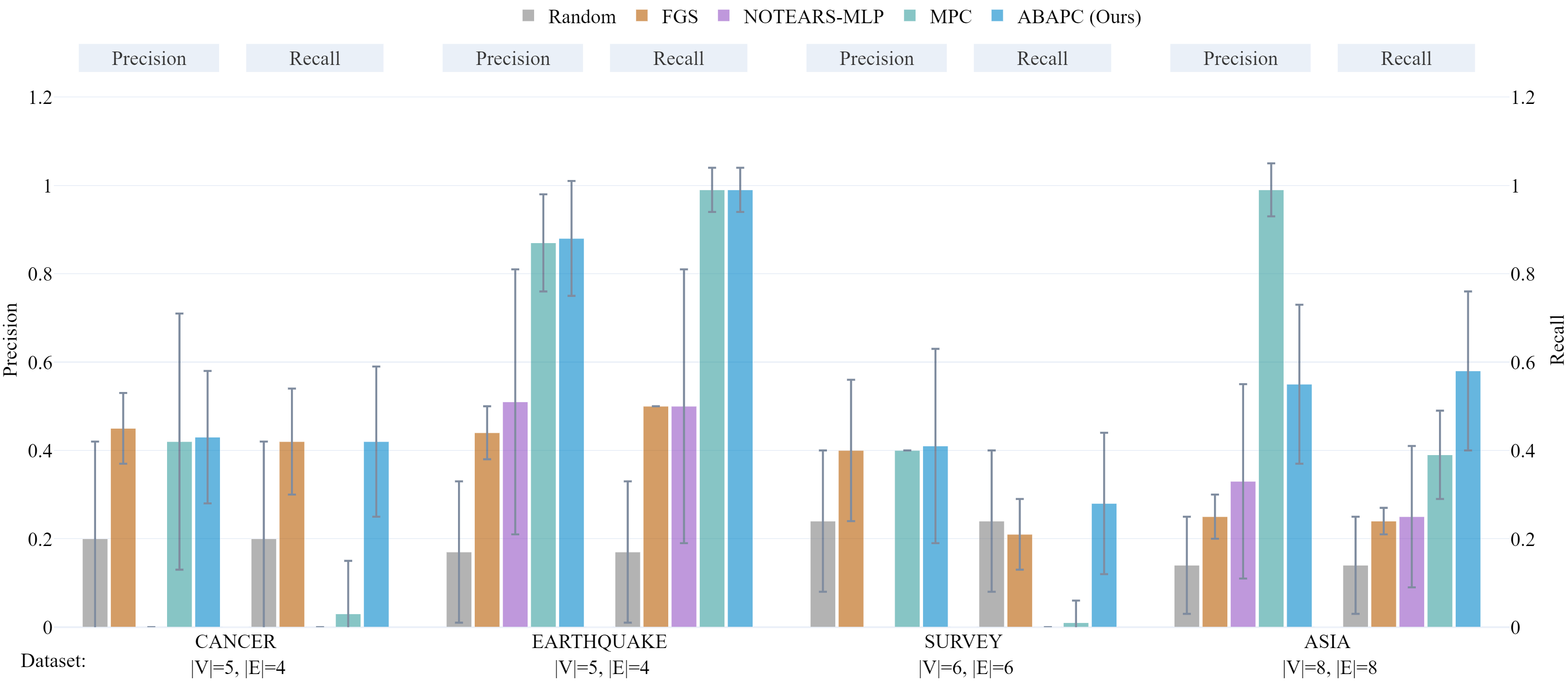}
    \caption{Precision (left y-axis) and Recall (right y-axis) for the estimated DAGs for four datasets in the \texttt{bnlearn} repository. Higher is better for both metrics.}
    \label{fig:pre_rec_DAG}
\end{figure*}

We report the evaluation of DAGs in Fig.~\ref{fig:SHD_SID_DAG} (NSID and NSHD) and Fig.~\ref{fig:pre_rec_DAG} (Precision and Recall). The plot of the F1 score is provided in the our repository both as image and interactive files.\footnote{\url{https://github.com/briziorusso/ArgCausalDisco/tree/public/results/figs}}According to NSHD, we can see that ABA-PC performs significantly better than all baselines on two out of the four datasets (Survey and Survey). For the Earthquake and Asia datasets ABA-PC is on par with MPC and significantly better than the other baselines. According to NSID, ABA-PC is not significantly different than MPC, ranking $1^{st}$ for the Earthquake and Asia datasets, and in line with all other baselines for the other two datsets. According to precision and recall, ABAPC is better than MPC for three out of the four datasets when evaluating on recall, while maintaining the same precision, again, three out of four times. For the Asia dataset ABAPC trades some precision for an significantly higher recall.

\subsection{Statistical Tests}\label{sec:stats}
Here we present details of the statistical tests used to measure the significance of the difference in the results presented in Fig.~\ref{fig:SID_cpdag} in the main text. In tables~\ref{tab:cancer_tests}, \ref{tab:earth_tests}, \ref{tab:survey_tests} and \ref{tab:asia_tests} we provide t-statistics and $p$-values for the Cancer, Earthquake, Survey and Asia datasets, respectively. In each table we present pairwise comparisons of means (shown in brakets together with standard deviations), for the best and worst case SID (High and Low, resp.) presented in Fig.~\ref{fig:SID_cpdag} of the main text.


\clearpage
\begin{table}[t]
    \caption{t-tests for difference in means for Cancer dataset. \\ Significance levels: 0 '***' 0.001 '**' 0.01 '*' 0.05 '.' 0.1 ' ' 1.}
    \label{tab:cancer_tests}
    \centering
    \begin{tabular}{rrl}
        \!\!\! Method (mean$\pm$std)  & \!\!\!t & \!\!\!p-value \\
        \hline
        \multicolumn{1}{l}{NSID (low)}&& \\
        \hline        \\[\dimexpr-\normalbaselineskip+2pt]
\!\!\!APC $(9.8\pm2.2)$ \!v\! FGS $(6.3\pm1.9)$ \!\!\!&\!\!\! 8.722 \!\!\!&\!\!\! 0.000*** \\
\!\!\!APC $(9.8\pm2.2)$ \!v\! \textbf{MPC $(3.9\pm1.8)$} \!\!\!&\!\!\! 14.833 \!\!\!&\!\!\! 0.000*** \\
\!\!\!APC $(9.8\pm2.2)$ \!v\! NT $(9.3\pm1.6)$ \!\!\!&\!\!\! 1.476 \!\!\!&\!\!\! 0.144 \\
\!\!\!APC $(9.8\pm2.2)$ \!v\! RND $(7.9\pm3.6)$ \!\!\!&\!\!\! 3.169 \!\!\!&\!\!\! 0.002** \\
\!\!\!FGS $(6.3\pm1.9)$ \!v\! MPC $(3.9\pm1.8)$ \!\!\!&\!\!\! 6.503 \!\!\!&\!\!\! 0.000*** \\
\!\!\!FGS $(6.3\pm1.9)$ \!v\! NT $(9.3\pm1.6)$ \!\!\!&\!\!\! -8.706 \!\!\!&\!\!\! 0.000*** \\
\!\!\!FGS $(6.3\pm1.9)$ \!v\! RND $(7.9\pm3.6)$ \!\!\!&\!\!\! -2.871 \!\!\!&\!\!\! 0.005** \\
\!\!\!MPC $(3.9\pm1.8)$ \!v\! NT $(9.3\pm1.6)$ \!\!\!&\!\!\! -16.024 \!\!\!&\!\!\! 0.000*** \\
\!\!\!MPC $(3.9\pm1.8)$ \!v\! RND $(7.9\pm3.6)$ \!\!\!&\!\!\! -7.078 \!\!\!&\!\!\! 0.000*** \\
\!\!\!NT $(9.3\pm1.6)$ \!v\! RND $(7.9\pm3.6)$ \!\!\!&\!\!\! 2.401 \!\!\!&\!\!\! 0.019* \\
        \\[\dimexpr-\normalbaselineskip+6pt]
        \hline \\[\dimexpr-\normalbaselineskip+2pt]
        \multicolumn{1}{l}{NSID (high)}&& \\
        \hline \\[\dimexpr-\normalbaselineskip+2pt]
\!\!\!APC $(11.0\pm2.1)$ \!v\! FGS $(16.3\pm1.9)$ \!\!\!&\!\!\! -13.231 \!\!\!&\!\!\! 0.000*** \\
\!\!\!APC $(11.0\pm2.1)$ \!v\! MPC $(13.9\pm1.8)$ \!\!\!&\!\!\! -7.315 \!\!\!&\!\!\! 0.000*** \\
\!\!\!APC $(11.0\pm2.1)$ \!v\! \textbf{NT $(10.2\pm0.4)$} \!\!\!&\!\!\! 2.794 \!\!\!&\!\!\! 0.007** \\
\!\!\!APC $(11.0\pm2.1)$ \!v\! RND $(14.0\pm3.0)$ \!\!\!&\!\!\! -5.801 \!\!\!&\!\!\! 0.000*** \\
\!\!\!FGS $(16.3\pm1.9)$ \!v\! MPC $(13.9\pm1.8)$ \!\!\!&\!\!\! 6.503 \!\!\!&\!\!\! 0.000*** \\
\!\!\!FGS $(16.3\pm1.9)$ \!v\! NT $(10.2\pm0.4)$ \!\!\!&\!\!\! 22.465 \!\!\!&\!\!\! 0.000*** \\
\!\!\!FGS $(16.3\pm1.9)$ \!v\! RND $(14.0\pm3.0)$ \!\!\!&\!\!\! 4.442 \!\!\!&\!\!\! 0.000*** \\
\!\!\!MPC $(13.9\pm1.8)$ \!v\! NT $(10.2\pm0.4)$ \!\!\!&\!\!\! 14.130 \!\!\!&\!\!\! 0.000*** \\
\!\!\!MPC $(13.9\pm1.8)$ \!v\! RND $(14.0\pm3.0)$ \!\!\!&\!\!\! -0.321 \!\!\!&\!\!\! 0.749 \\
\!\!\!NT $(10.2\pm0.4)$ \!v\! RND $(14.0\pm3.0)$ \!\!\!&\!\!\! -8.934 \!\!\!&\!\!\! 0.000*** \\
    \end{tabular}
\end{table}

\begin{table}[b]
    \caption{t-tests for difference in means for Earthquake dataset. \\ Significance levels: 0 '***' 0.001 '**' 0.01 '*' 0.05 '.' 0.1 ' ' 1.}
    \label{tab:earth_tests}
    \centering
    \begin{tabular}{rrl}
        \!\!\! Method (mean$\pm$std)  & \!\!\!t & \!\!\!p-value \\
        \hline
        \multicolumn{1}{l}{NSID (low)}&& \\
        \hline        \\[\dimexpr-\normalbaselineskip+2pt]
\!\!\!\textbf{APC $(0.0\pm0.0)$} \!v\! FGS $(4.6\pm1.4)$ \!\!\!&\!\!\! -23.742 \!\!\!&\!\!\! 0.000*** \\
\!\!\!APC $(0.0\pm0.0)$ \!v\! MPC $(5.0\pm0.0)$ \!\!\!&\!\!\! -inf \!\!\!&\!\!\! 0.000*** \\
\!\!\!APC $(0.0\pm0.0)$ \!v\! NT $(4.6\pm3.3)$ \!\!\!&\!\!\! -9.741 \!\!\!&\!\!\! 0.000*** \\
\!\!\!APC $(0.0\pm0.0)$ \!v\! RND $(7.9\pm2.9)$ \!\!\!&\!\!\! -19.347 \!\!\!&\!\!\! 0.000*** \\
\!\!\!FGS $(4.6\pm1.4)$ \!v\! MPC $(5.0\pm0.0)$ \!\!\!&\!\!\! -2.065 \!\!\!&\!\!\! 0.044* \\
\!\!\!FGS $(4.6\pm1.4)$ \!v\! NT $(4.6\pm3.3)$ \!\!\!&\!\!\! 0.079 \!\!\!&\!\!\! 0.937 \\
\!\!\!FGS $(4.6\pm1.4)$ \!v\! RND $(7.9\pm2.9)$ \!\!\!&\!\!\! -7.272 \!\!\!&\!\!\! 0.000*** \\
\!\!\!MPC $(5.0\pm0.0)$ \!v\! NT $(4.6\pm3.3)$ \!\!\!&\!\!\! 0.940 \!\!\!&\!\!\! 0.352 \\
\!\!\!MPC $(5.0\pm0.0)$ \!v\! RND $(7.9\pm2.9)$ \!\!\!&\!\!\! -7.071 \!\!\!&\!\!\! 0.000*** \\
\!\!\!NT $(4.6\pm3.3)$ \!v\! RND $(7.9\pm2.9)$ \!\!\!&\!\!\! -5.351 \!\!\!&\!\!\! 0.000*** \\
        \\[\dimexpr-\normalbaselineskip+6pt]
        \hline \\[\dimexpr-\normalbaselineskip+2pt]
        \multicolumn{1}{l}{NSID (high)}&& \\
        \hline \\[\dimexpr-\normalbaselineskip+2pt]
\!\!\!\textbf{APC $(9.3\pm4.5)$} \!v\! FGS $(15.0\pm0.5)$ \!\!\!&\!\!\! -8.796 \!\!\!&\!\!\! 0.000*** \\
\!\!\!APC $(9.3\pm4.5)$ \!v\! MPC $(15.0\pm0.0)$ \!\!\!&\!\!\! -8.847 \!\!\!&\!\!\! 0.000*** \\
\!\!\!APC $(9.3\pm4.5)$ \!v\! NT $(13.5\pm4.0)$ \!\!\!&\!\!\! -4.812 \!\!\!&\!\!\! 0.000*** \\
\!\!\!APC $(9.3\pm4.5)$ \!v\! RND $(13.9\pm3.5)$ \!\!\!&\!\!\! -5.649 \!\!\!&\!\!\! 0.000*** \\
\!\!\!FGS $(15.0\pm0.5)$ \!v\! MPC $(15.0\pm0.0)$ \!\!\!&\!\!\! 0.000 \!\!\!&\!\!\! 1.000 \\
\!\!\!FGS $(15.0\pm0.5)$ \!v\! NT $(13.5\pm4.0)$ \!\!\!&\!\!\! 2.669 \!\!\!&\!\!\! 0.010* \\
\!\!\!FGS $(15.0\pm0.5)$ \!v\! RND $(13.9\pm3.5)$ \!\!\!&\!\!\! 2.266 \!\!\!&\!\!\! 0.028* \\
\!\!\!MPC $(15.0\pm0.0)$ \!v\! NT $(13.5\pm4.0)$ \!\!\!&\!\!\! 2.689 \!\!\!&\!\!\! 0.010** \\
\!\!\!MPC $(15.0\pm0.0)$ \!v\! RND $(13.9\pm3.5)$ \!\!\!&\!\!\! 2.289 \!\!\!&\!\!\! 0.026* \\
\!\!\!NT $(13.5\pm4.0)$ \!v\! RND $(13.9\pm3.5)$ \!\!\!&\!\!\! -0.558 \!\!\!&\!\!\! 0.578 \\
    \end{tabular}
\end{table}

\begin{table}[t]
    \caption{t-tests for difference in means for Survey dataset. \\ Significance levels: 0 '***' 0.001 '**' 0.01 '*' 0.05 '.' 0.1 ' ' 1.}
    \label{tab:survey_tests}
    \centering
    \begin{tabular}{rrl}
        \!\!\! Method (mean$\pm$std)  & \!\!\!t & \!\!\!p-value \\
        \hline
        \multicolumn{1}{l}{NSID (low)}&& \\
        \hline        \\[\dimexpr-\normalbaselineskip+2pt]
\!\!\!APC $(14.2\pm3.6)$ \!v\! FGS $(13.7\pm3.2)$ \!\!\!&\!\!\! 0.819 \!\!\!&\!\!\! 0.415 \\
\!\!\!APC $(14.2\pm3.6)$ \!v\! \textbf{MPC $(5.1\pm3.4)$} \!\!\!&\!\!\! 12.854 \!\!\!&\!\!\! 0.000*** \\
\!\!\!APC $(14.2\pm3.6)$ \!v\! NT $(17.0\pm0.0)$ \!\!\!&\!\!\! -5.386 \!\!\!&\!\!\! 0.000*** \\
\!\!\!APC $(14.2\pm3.6)$ \!v\! RND $(13.1\pm4.2)$ \!\!\!&\!\!\! 1.405 \!\!\!&\!\!\! 0.163 \\
\!\!\!FGS $(13.7\pm3.2)$ \!v\! MPC $(5.1\pm3.4)$ \!\!\!&\!\!\! 12.940 \!\!\!&\!\!\! 0.000*** \\
\!\!\!FGS $(13.7\pm3.2)$ \!v\! NT $(17.0\pm0.0)$ \!\!\!&\!\!\! -7.450 \!\!\!&\!\!\! 0.000*** \\
\!\!\!FGS $(13.7\pm3.2)$ \!v\! RND $(13.1\pm4.2)$ \!\!\!&\!\!\! 0.730 \!\!\!&\!\!\! 0.467 \\
\!\!\!MPC $(5.1\pm3.4)$ \!v\! NT $(17.0\pm0.0)$ \!\!\!&\!\!\! -24.593 \!\!\!&\!\!\! 0.000*** \\
\!\!\!MPC $(5.1\pm3.4)$ \!v\! RND $(13.1\pm4.2)$ \!\!\!&\!\!\! -10.490 \!\!\!&\!\!\! 0.000*** \\
\!\!\!NT $(17.0\pm0.0)$ \!v\! RND $(13.1\pm4.2)$ \!\!\!&\!\!\! 6.595 \!\!\!&\!\!\! 0.000*** \\
        \\[\dimexpr-\normalbaselineskip+6pt]
        \hline \\[\dimexpr-\normalbaselineskip+2pt]
        \multicolumn{1}{l}{NSID (high)}&& \\
        \hline \\[\dimexpr-\normalbaselineskip+2pt]
\!\!\!\textbf{APC $(16.4\pm2.8)$} \!v\! FGS $(20.7\pm3.5)$ \!\!\!&\!\!\! -6.764 \!\!\!&\!\!\! 0.000*** \\
\!\!\!APC $(16.4\pm2.8)$ \!v\! MPC $(20.1\pm4.4)$ \!\!\!&\!\!\! -5.065 \!\!\!&\!\!\! 0.000*** \\
\!\!\!APC $(16.4\pm2.8)$ \!v\! NT $(17.0\pm0.0)$ \!\!\!&\!\!\! -1.481 \!\!\!&\!\!\! 0.145 \\
\!\!\!APC $(16.4\pm2.8)$ \!v\! RND $(22.4\pm4.4)$ \!\!\!&\!\!\! -8.092 \!\!\!&\!\!\! 0.000*** \\
\!\!\!FGS $(20.7\pm3.5)$ \!v\! MPC $(20.1\pm4.4)$ \!\!\!&\!\!\! 0.686 \!\!\!&\!\!\! 0.494 \\
\!\!\!FGS $(20.7\pm3.5)$ \!v\! NT $(17.0\pm0.0)$ \!\!\!&\!\!\! 7.480 \!\!\!&\!\!\! 0.000*** \\
\!\!\!FGS $(20.7\pm3.5)$ \!v\! RND $(22.4\pm4.4)$ \!\!\!&\!\!\! -2.151 \!\!\!&\!\!\! 0.034* \\
\!\!\!MPC $(20.1\pm4.4)$ \!v\! NT $(17.0\pm0.0)$ \!\!\!&\!\!\! 5.060 \!\!\!&\!\!\! 0.000*** \\
\!\!\!MPC $(20.1\pm4.4)$ \!v\! RND $(22.4\pm4.4)$ \!\!\!&\!\!\! -2.560 \!\!\!&\!\!\! 0.012* \\
\!\!\!NT $(17.0\pm0.0)$ \!v\! RND $(22.4\pm4.4)$ \!\!\!&\!\!\! -8.633 \!\!\!&\!\!\! 0.000*** \\
    \end{tabular}
\end{table}

\begin{table}[b]
    \caption{t-tests for difference in means for Asia dataset. \\ Significance levels: 0 '***' 0.001 '**' 0.01 '*' 0.05 '.' 0.1 ' ' 1.}
    \label{tab:asia_tests}
    \centering
    \begin{tabular}{rrl}
        \!\!\! Method (mean$\pm$std)  & \!\!\!t & \!\!\!p-value \\
        \hline
        \multicolumn{1}{l}{SID (low)}&& \\
        \hline        \\[\dimexpr-\normalbaselineskip+2pt]
\!\!\!\textbf{APC $(11.7\pm6.8)$} \!v\! FGS $(32.3\pm5.0)$ \!\!\!&\!\!\! -17.164 \!\!\!&\!\!\! 0.000*** \\
\!\!\!APC $(11.7\pm6.8)$ \!v\! MPC $(15.2\pm5.6)$ \!\!\!&\!\!\! -2.828 \!\!\!&\!\!\! 0.006** \\
\!\!\!APC $(11.7\pm6.8)$ \!v\! NT $(16.5\pm4.2)$ \!\!\!&\!\!\! -4.243 \!\!\!&\!\!\! 0.000*** \\
\!\!\!APC $(11.7\pm6.8)$ \!v\! RND $(25.0\pm7.0)$ \!\!\!&\!\!\! -9.622 \!\!\!&\!\!\! 0.000*** \\
\!\!\!FGS $(32.3\pm5.0)$ \!v\! MPC $(15.2\pm5.6)$ \!\!\!&\!\!\! 15.960 \!\!\!&\!\!\! 0.000*** \\
\!\!\!FGS $(32.3\pm5.0)$ \!v\! NT $(16.5\pm4.2)$ \!\!\!&\!\!\! 16.895 \!\!\!&\!\!\! 0.000*** \\
\!\!\!FGS $(32.3\pm5.0)$ \!v\! RND $(25.0\pm7.0)$ \!\!\!&\!\!\! 5.970 \!\!\!&\!\!\! 0.000*** \\
\!\!\!MPC $(15.2\pm5.6)$ \!v\! NT $(16.5\pm4.2)$ \!\!\!&\!\!\! -1.290 \!\!\!&\!\!\! 0.200 \\
\!\!\!MPC $(15.2\pm5.6)$ \!v\! RND $(25.0\pm7.0)$ \!\!\!&\!\!\! -7.690 \!\!\!&\!\!\! 0.000*** \\
\!\!\!NT $(16.5\pm4.2)$ \!v\! RND $(25.0\pm7.0)$ \!\!\!&\!\!\! -7.322 \!\!\!&\!\!\! 0.000*** \\
        \\[\dimexpr-\normalbaselineskip+6pt]
        \hline \\[\dimexpr-\normalbaselineskip+2pt]
        \multicolumn{1}{l}{SID (high)}&& \\
        \hline \\[\dimexpr-\normalbaselineskip+2pt]
\!\!\!\textbf{APC $(33.5\pm7.9)$} \!v\! FGS $(41.7\pm2.8)$ \!\!\!&\!\!\! -6.883 \!\!\!&\!\!\! 0.000*** \\
\!\!\!APC $(33.5\pm7.9)$ \!v\! MPC $(41.0\pm3.8)$ \!\!\!&\!\!\! -6.046 \!\!\!&\!\!\! 0.000*** \\
\!\!\!APC $(33.5\pm7.9)$ \!v\! NT $(41.1\pm5.0)$ \!\!\!&\!\!\! -5.739 \!\!\!&\!\!\! 0.000*** \\
\!\!\!APC $(33.5\pm7.9)$ \!v\! RND $(37.2\pm6.1)$ \!\!\!&\!\!\! -2.633 \!\!\!&\!\!\! 0.010** \\
\!\!\!FGS $(41.7\pm2.8)$ \!v\! MPC $(41.0\pm3.8)$ \!\!\!&\!\!\! 1.023 \!\!\!&\!\!\! 0.309 \\
\!\!\!FGS $(41.7\pm2.8)$ \!v\! NT $(41.1\pm5.0)$ \!\!\!&\!\!\! 0.745 \!\!\!&\!\!\! 0.458 \\
\!\!\!FGS $(41.7\pm2.8)$ \!v\! RND $(37.2\pm6.1)$ \!\!\!&\!\!\! 4.702 \!\!\!&\!\!\! 0.000*** \\
\!\!\!MPC $(41.0\pm3.8)$ \!v\! NT $(41.1\pm5.0)$ \!\!\!&\!\!\! -0.091 \!\!\!&\!\!\! 0.928 \\
\!\!\!MPC $(41.0\pm3.8)$ \!v\! RND $(37.2\pm6.1)$ \!\!\!&\!\!\! 3.732 \!\!\!&\!\!\! 0.000*** \\
\!\!\!NT $(41.1\pm5.0)$ \!v\! RND $(37.2\pm6.1)$ \!\!\!&\!\!\! 3.478 \!\!\!&\!\!\! 0.001*** \\
    \end{tabular}
\end{table}

\end{document}